%% file: main.tex
\documentclass[sigconf]{acmart}
\renewcommand\footnotetextcopyrightpermission[1]{}
\AtBeginDocument{%
  }

\usepackage{multirow}
\usepackage{bm}
\usepackage{amsmath}
\usepackage{algorithmic}
\usepackage[linesnumbered,ruled,vlined]{algorithm2e}
\SetKwInput{KwInput}{Input}                
\SetKwInput{KwOutput}{Output}              
\SetKwInput{KwParameter}{Parameter}        
\usepackage{listings}
\usepackage{subfigure}
\usepackage{graphicx}
\usepackage{blindtext}

\usepackage{url} 

\makeatletter
\def\@copyrightspace{\relax}
\makeatother

\begin{document}



\title{\our: Learning Causal Structure for Marketing Mix Modeling}
\author{Chang Gong}
\affiliation{
    \institution{
    Institute of Computing Technology, Chinese Academy of Sciences\\
    University of Chinese Academy of Sciences\\
    \country{China}
    }
}
\email{gongchang21z@ict.ac.cn}

\author{Di Yao}
\affiliation{
    \institution{
    Institute of Computing Technology, Chinese Academy of Sciences\\
    \country{China}
    }
}
\email{yaodi@ict.ac.cn}
\authornote{Corresponding authors.}

\author{Lei Zhang}
\affiliation{
    \institution{
    Independent Researcher\\
    \country{China}
    }
}
\email{zhanglei342@gmail.com}

\author{Sheng Chen}
\affiliation{
    \institution{
    Independent Researcher\\
    \country{China}
    }
}
\email{chensheng.cn@gmail.com}

\author{Wenbin Li}
\affiliation{
    \institution{
    Institute of Computing Technology, Chinese Academy of Sciences\\
    University of Chinese Academy of Sciences\\
    \country{China}
    }
}
\email{liwenbin20z@ict.ac.cn}

\author{Yueyang Su}
\affiliation{
    \institution{
    Institute of Computing Technology, Chinese Academy of Sciences\\
    University of Chinese Academy of Sciences\\
    \country{China}
    }
}
\email{suyueyang19b@ict.ac.cn}

\author{Jingping Bi}
\affiliation{
    \institution{
    Institute of Computing Technology, Chinese Academy of Sciences\\
    \country{China}
    }
}
\email{bjp@ict.ac.cn}
\authornotemark[1]

\renewcommand{\shortauthors}{Chang Gong et al.}
\newcommand{\our}{\texttt{CausalMMM}\xspace}

\newcommand{\ourfu}{\textsc{CM-full}\xspace}
\newcommand{\ourrw}{\textsc{CM-rw}\xspace}
\newcommand{\oursup}{\textsc{CM-supervised}\xspace}
\newcommand{\ourmar}{\textsc{CM-markov}\xspace}

\newcommand{\ie}{\emph{i.e.}\xspace} 
\newcommand{\etal}{\emph{et~al.}\xspace} 
\newcommand{\etc}{\emph{etc.}\xspace} 
\newcommand{\eg}{\emph{e.g.}\xspace} 
\newcommand{\define}[3]{\vspace{1ex}\noindent{ \textbf{\textsc{Definition {#1}}} (#2): \emph{#3}\vspace{1ex}}
}
\newcommand{\hypothe}[3]{\vspace{1ex}\noindent{ \textbf{\textsc{Hypothesis {#1}}} (#2): \emph{#3}\vspace{1ex}}
}

\newcommand{\claim}[3]{\vspace{1ex}\noindent{ \textbf{\textsc{Claim {#1}}} (#2): \emph{#3}\vspace{1ex}}
}

\newcommand\tab[1][0.5cm]{\hspace*{#1}}
\newcommand{\red}[1]{\textcolor{red}{#1}}
\newcommand{\blue}[1]{\textcolor{blue}{#1}}


\input{sections/abstract.tex}




\maketitle
\input{sections/introduction.tex}

\input{sections/preliminary.tex}
\input{sections/method.tex}

\input{sections/experiment.tex}

\input{sections/related.tex}

\input{sections/conclusion.tex}

\input{sections/acknowledgements.tex}

\bibliographystyle{ACM-Reference-Format}
\bibliography{references}

\input{sections/ethical_consideration}

\clearpage
\appendix
\input{sections/appendix.tex}

\end{document}

%% file: sections/abstract.tex
\begin{abstract}
In online advertising, marketing mix modeling (MMM) is employed to predict the gross merchandise volume (GMV) of brand shops and help decision-makers to adjust the budget allocation of various advertising channels.
Traditional MMM methods leveraging regression techniques can only capture linear relationships and fail in handling the complexity of marketing. 
Although some efforts try to encode the causal structures for better prediction, they have the strict restriction that causal structures are prior-known and unchangeable. 
In this paper, we define a new causal MMM problem that automatically discovers the interpretable causal structures from data and yields better GMV predictions.
To achieve causal MMM, two essential challenges should be addressed: (1) Causal Heterogeneity. The causal structures of different kinds of shops vary a lot. (2) Marketing Response Patterns. Various marketing response patterns \ie, carryover effect and shape effect, have been validated in practice. We argue that causal MMM needs dynamically discover specific causal structures for different shops and the predictions should comply with the prior known marketing response patterns.
Thus, we propose \our that integrates Granger causality in a variational inference framework to measure the causal relationships between different channels and predict the GMV with the regularization of both temporal and saturation marketing response patterns.
Extensive experiments show that \our can not only achieve superior performance of causal structure learning on synthetic datasets with improvements of $5.7\%\sim 7.1\%$, but also enhance the GMV prediction results on a representative E-commerce platform.

\end{abstract}

%% file: sections/introduction.tex
\section{Introduction}

\begin{figure}
    \centering
	\includegraphics[width=0.5\textwidth]{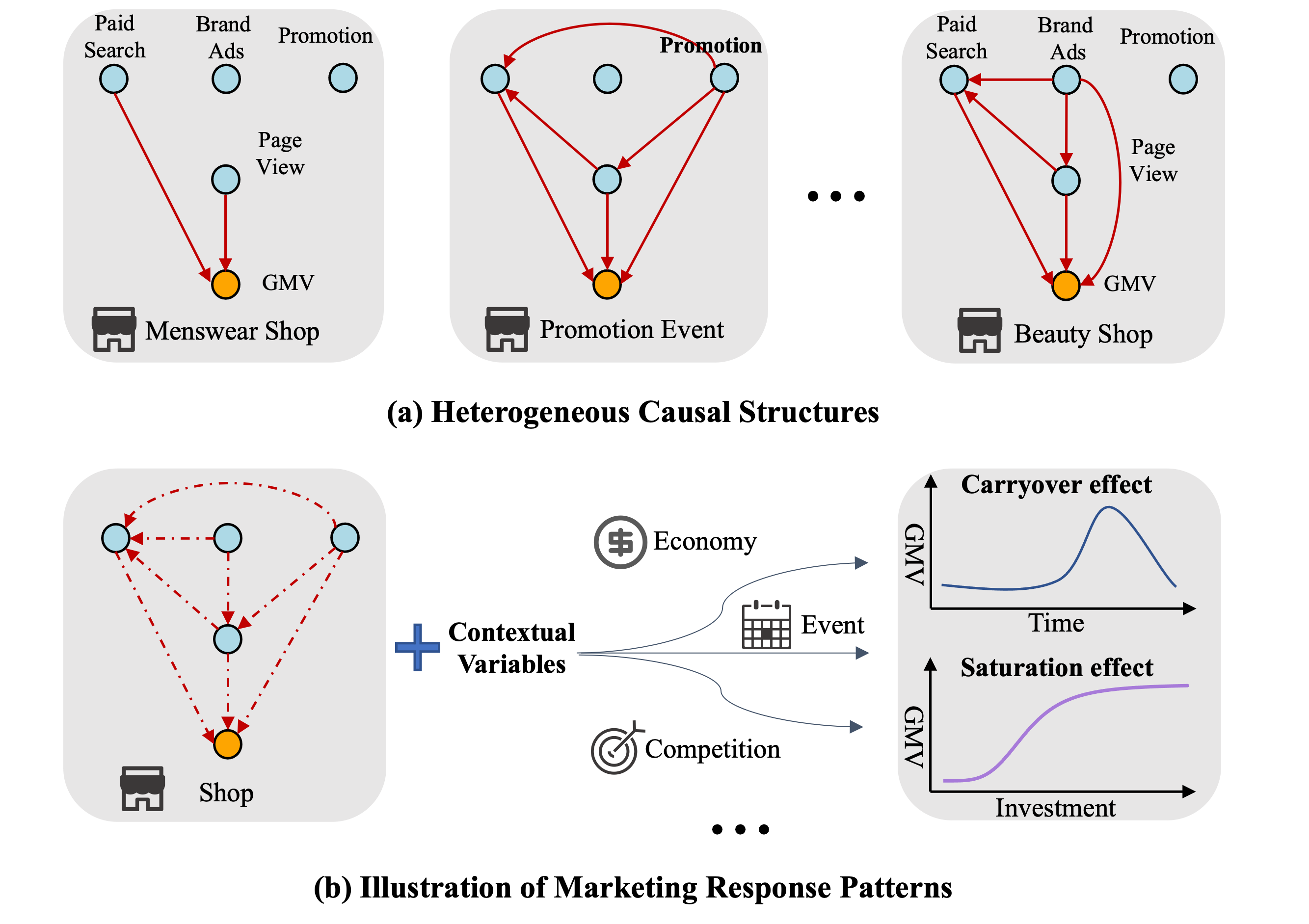} 
	\vspace{-5ex} 
	\caption{The motivation of \our. (a) shows heterogeneous causal structures in MMM where nodes in different colors denote channel and target variables, respectively. (b) illustrates the saturation curve of marketing response determined by contextual factors, such as economy, events, \etc.}
	\vspace{-5.5ex}
	\label{fig:MMM_motivation}
\end{figure}

Online advertising platforms enable advertisers to launch their ads on various marketing channels, \eg, paid search, feed stream, and \etc How to allocate advertising budget on different channels to maximize the GMV remains a critical yet unsettled issue. Marketing mix modeling, short for MMM, try to address this issue as a prediction problem. Taking channel costs and shop attributes as input, MMM methods aim to predict the GMV in the future and estimate the effectiveness of costs in sales. The results can shed light on advertising cost control and performance growth. Therefore, MMM is of vital importance for both advertisers and online platforms.  




In the last decade, MMM has become a popular problem and attracted many researchers. Because advertisers need reasons for making decisions, interpretability is essential for MMM and simple models are preferred to be used as the predictor. The majority of existing works can be divided into two categories, regression-based methods~\cite{comp_survey/wigren2019marketing, MMM_google/wang2017hierarchical} and causal-based methods~\cite{MMM_google/chen2018bias}. Methods in the first category~\cite{adkdd/ng2021bayesian, JAS/chen2021hierarchical} predict the GMV with the regression of different channel costs and employ the coefficients as evidence of interpretation. The implicit assumption is that the influences of all channels are independent and would be direct causes of GMV. This assumption can be easily violated in practice, \eg, ads in display channels increasing users' awareness will promote search channels to deals. Regression-based methods ignore this effect leading to their poor performance. For causal-based methods, an ad-hoc causal structure of channels is pre-defined as the prior knowledge to build the prediction model. However, for different shops, the influences of channels are also different. Without considering the heterogeneity, this kind of method can not be used to guide decision-making accurately. Moreover, current MMM methods predict the GMV with linear models, which can not capture the complexity of marketing. 

Witnessing the drawbacks of current works, we argue that MMM should be able to dynamically discover the causal structures of different channels and predict the GMV in small error. Thus, a new task causal MMM is defined in this paper. Compared with traditional MMM, causal MMM integrates the causal structure learning of channels into the GMV prediction and interprets channels' influences with the effectiveness of causal relations. Benefitting from explicitly discovered causal structure, the model can be easily generalized on long-term predictions and the advertisers gain new insights.

Nevertheless, it's often non-trivial to achieve causal MMM. The reasons are two folds:
(1) \textbf{Causal Heterogeneity}. The causal structures and dynamics of channels vary according to the characters of shops and in different periods. As shown in Figure \ref{fig:MMM_motivation}(a), the underlying causal drivers of sales volume and the curves of marketing response are often heterogeneous across different brands' profiles and marketing events. The causal effect between brand ads and page views in beauty shops are decisive compared with menswear shops. This heterogeneity of causal structures is further challenging to discover under the circumstance of data imbalance.
(2) \textbf{Marketing Response Patterns}. In previous studies~\cite{MMM_46001/jin2017bayesian, bg_adeffect/martinez2022distributed, bg_adeffect/hanssens2003market}, several important patterns in advertising response, such as carryover effect and saturation effect, have been validated. As illustrated in Figure \ref{fig:MMM_motivation}, the influence of advertising investment has time decay and would be saturated with the increase of investment. In addition to the causal structure, GMV curves are also affected by some contextual variables such as economy and competition. Thus, the predictor in causal MMM should be designed carefully to satisfy marketing response patterns.

To address these challenges, we propose a novel marketing mix model, called \underline{\textit{\our}}, which tackles both causal heterogeneity and marketing patterns simultaneously and achieves causal MMM. Overall, \our is a graph variational autoencoder-based method, which consists of two key modules: the causal relational encoder and the marketing response decoder. Although the causal structures are heterogeneous, the causal mechanisms are shared in different shops. Thus, in causal relational encoder, we encode the historic data of shops to generate the specified causal structures with Gumbel softmax sampling. Based on the causal structure, the marketing response decoder is designed to satisfy the priors of marketing response patterns and achieve good prediction performance. The sequential models and S-curve transformation are integrated into the decoder to capture the carryover and saturation effects respectively. For optimization, \our directly fits the historical data with the variational inference loss and learns the parameters in an end-to-end manner. Moreover, \our has a theoretical guarantee that the obtained causal structures are Granger causality. In a nutshell, the contributions of this paper are summarized as follows: 
\begin{itemize}
    \item We define a new task, called causal MMM, which discovers the causal relations among channels and marketing targets. Compared with traditional MMM, the new task provides more insights for decision-makers in online advertising.
    \item We propose a novel method \our, which is the first neural network-based solution for MMM problem. It is provable for learning heterogeneous Granger causal structures of different shops and capable of modeling patterns in marketing response.
    \item Extensive experiments on a synthetic dataset and a real-world dataset from an E-commerce platform demonstrate \our's superiority. It not only shows its effectiveness on causal structure learning with improvements of $5.7\%\sim 7.1\%$ on a synthetic dataset, but also can derive comparative prediction results on a real-world dataset.  
\end{itemize}

%% file: sections/preliminary.tex
\section{Preliminary}

\subsection{Causality Basics}
Granger causality~\cite{granger1969investigating} is a widely used method for inferring causal relationships from observational temporal data, based on the assumption that causes precede their effects.
It indicates that if knowing past elements of X can enhance the prediction of future Y, then X ``Granger causes'' Y.
Granger causality was initially defined for linear relationships.
To extend to more general cases, we follow the formal definition of~\cite{MTS/Granger/pamiNGC22, MTS/Granger/iclr20_esru, MTS/Granger/iclr21_GVAR_MarcinkevicsV, Discussion/NewForm/ACD_LoweMSW22} for non-linear Granger causality. 

\define{1}{Non-Linear Granger Causality}{Given $d$ time-series $\mathbf{X}=(\mathbf{x}_{1}^{1:T}, ...,  \mathbf{x}_{d}^{1:T})$ across $T$ time points and a non-linear function $g_j$,
$$ \mathbf{x}_{j}^{t+1} = g_j ( \mathbf{x}_{1}^{1:t}, ...,  \mathbf{x}_{d}^{1:t} ) + \boldsymbol{\epsilon}_{j}^{t+1} $$       
where $\boldsymbol{\epsilon}_{j}^{t+1}$ denotes independent noise. Time-series $i$ Granger causes $j$, if $g_j$ depends on $\mathbf{x}_{i}^{1:t}$, i.e.}
$$\exists \mathbf{x}^{1:t}_i   \neq   \mathbf{x'}^{1:t}_i, \   g_j ( \mathbf{x}^{1:t}_1,..., \mathbf{x}^{1:t}_i, ..., \mathbf{x}^{1:t}_d )   \neq  g_j ( \mathbf{x}^{1:t}_1,..., \mathbf{x'}^{1:t}_i, ..., \mathbf{x}^{1:t}_d ).  $$

Given the formal definition, Granger causality can be summarized by a directed graph $\mathcal{G}=\{\mathcal{V}, \mathcal{E} \} $, referred to as summary graph~\cite{intro/ts_surveys/AssaadDG22}. 
Here, $\mathcal{V}$ is the set of vertices corresponding to variables, and $\mathcal{E} = \big\{ (i,j) : \mathbf{x}_i \to \mathbf{x}_j   \big\}$ is the set of edges corresponding to Granger causal relations.   
Let $\mathbf{{A}}$ denote the adjacency matrix of $\mathcal{G}$, the causal structure learning, or causal discovery, problem is then to estimate $\mathbf{A}$ from observational temporal data.

\begin{table}[t]
    \centering
    \caption{The primary notations in this paper.}
    \label{tab:notation_table}
    \small
    \begin{tabular}{l|l}  
    \toprule
    \multicolumn{2}{c}{\textbf{Marketing Mix Data}}  \\
    \midrule

$N $ & Number of shops or brands \\
$\mathcal{D}$ & Marketing mix dataset \\
$\mathbf{X}$ & The advertising spend records of $d$ channels in $T$ days \\
$\mathbf{y}^{1:T}   $ & Historical values of marketing target, \eg, GMV \\
$\mathbf{C}  $ &  Vector of contextual variable \\ 
    \midrule
    \multicolumn{2}{c}{\textbf{Graph Structures}}  \\
    \midrule

$\mathcal{G}=(\mathcal{V}, \mathcal{E})$ & Causal structure $\mathcal{G}$ with set of nodes $\mathcal{V}$ and set of edges $\mathcal{E}$ \\
$v_i $ &  A node $v_i \in \mathcal{V}$ \\
$e_{ij}$ & A directed edge $e_{ij} \in \mathcal{E}$ from $v_i$ to $v_j$ \\

    \midrule
    
    \multicolumn{2}{c}{\textbf{Model Architecture}}  \\
    \midrule
$q_{\tiny\phi},p_{\tiny\theta}$ & The encoder and the decoder, respectively \\
$f_{\mathrm{enc}}, f_{\mathrm{dec}}$ & Synonyms for the encoder and the decoder, respectively \\
$f_{\mathrm{vertex}}, f_{\mathrm{edge}} $   & Node-specific and edge-specific neural networks \\ 
$f_{\mathrm{seq}}, f_{\mathrm{pre}}$ & The sequence model and the prediction model in decoder \\
$f_{\alpha}, f_{\gamma} $ &  Neural networks control the shape of saturation curves \\
$\mathbf{h}_i^l,\mathbf{h}_{ij}^l $ & The embedding of node $v_i$ and edge $e_ij$ in the $l$-th layer\\
 & of the GNNs in the encoder, respectively \\
$\tilde{\mathbf{h}}_{j}^{t}, \tilde{\mathbf{h}}_{j}^{t}$ & The embedding of node $v_i$ and edge $e_ij$ at $t$ in the decoder \\ 
$\mathbf{z}$ & The latent causal structure matrix \\
${z}_{ij}$ & Value representing the likelihood of of edge $e_{ij}$ \\
$\mathrm{MSG}_j^t $ & The aggregate causal information for node $v_j$ at $t$ \\
$\boldsymbol{\mu}_{j}^{t+1}$ & The mean value of prediction for node $v_j$ at $t+1$\\
$\lambda$ & The penalty factor of the structural prior \\
$\tau$ & The temperature parameter \\
\bottomrule
\end{tabular} 
\vspace{-4ex}
\end{table}

\subsection{Problem Definition}
In this paper, we study the problem of causal MMM based on a marketing mix dataset, which aims to build a model that simultaneously (1) infers causal structure among marketing variables for each shop; (2) predicts the marketing target complying with the prior known marketing response patterns. 
The marketing mix dataset is defined as: 

\define{2}{Marketing Mix Dataset}{A marketing mix dataset $\mathcal{D}$ consists of marketing records from $N$ shops. For the $n$-th shop, its record can be formulated as a triplet, i.e., $(\mathbf{X}_n,  \mathbf{y}_n,  \mathbf{C}_n)$. $\mathbf{X_n}=(\mathbf{x}_{n,1}^{1:T}, ...,  \mathbf{x}_{n,d}^{1:T}) \in \mathbb{R}^{T \times d}$ is a $d$-variate time series representing the advertising spends on $d$ channels. $\mathbf{y}_n^T \in \mathbb{R}^T$ is the $n$-th shop's marketing target value, e.g., GMV. $\mathbf{C}_n$ represents the vector of contextual variables, including economy indices, event identity, and so on.
} 

The causal MMM problem is based on the premise that there exists a causal graph $\mathcal{G}_n = \{ \mathcal{V}_n, \mathcal{E}_n\}$ underlying the marketing process of each shop $n$, where $\mathcal{V}_n$ contains advertising spends $\mathbf{X}_n$ and marketing target $\mathbf{y}_n$.
Therefore, the goal of causal MMM problem is formulated as: given a marketing dataset $\mathcal{D}$, (1) infer causal structures $\{\mathcal{G}_1,...\ ,\mathcal{G}_N\}$, and (2) predict marketing target $\{\mathbf{y}_1^{T+1:T+M}, ...\ ,$ $\mathbf{y}_N^{T+1:T+M}  \}$ for each shop.

\subsection{Method Overview}
\begin{figure}
    \centering
        \label{fig:overview}
	\includegraphics[width=0.48\textwidth]{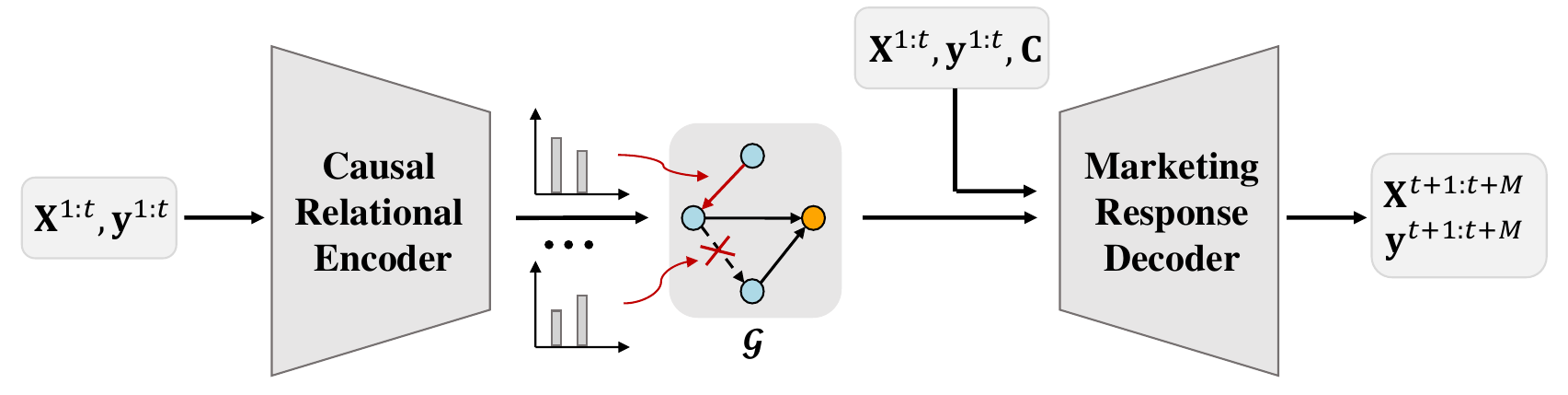} %
	\vspace{-5ex} 
	\caption{The overview of \our. The causal relational encoder predicts causal structures between marketing variables $\mathbf{X, y}$. The marketing response decoder learns to predict marketing variables given their past observations and contextual variables. This framework enables us to extract heterogeneous causal structures and learn marketing responses, simultaneously.}
	\vspace{-4ex}
	\label{fig:MMM_overview}
\end{figure}

As shown in Figure~\ref{fig:MMM_overview}, the proposed \our contains two concise parts, \ie, an encoder for causal structure learning, and a decoder for marketing response modeling. 
In the encoder, we aim to predict edges in the causal graph given observed marketing mix data so that heterogeneous causal structures across different shops can be mined. 
In the decoder, the marketing response is modeled under the inferred causal structures with the regularization of both temporal and saturation patterns. 
The temporal marketing response module conducts message passing on the inferred causal structure by taking the historical hidden states. 
The saturation marketing response module fits S-curve in an explicit way by leveraging learnable inflexion points and curve factors.
After that, we can systematically learn causal structure among heterogeneous marketing data and model advertising responses explicitly.
The reason why \our able to find Granger causal structures and the model complexity are analyzed respectively.

%% file: sections/method.tex
\section{Methodology}

\subsection{Causal Relational Encoder}
\label{sec:M2}
The causal relational encoder aims to infer the likelihood of causal relation $z_{ij}$ based on the historical advertising spend $(\mathbf{x}_{1}^{1:T}, ...,  \mathbf{x}_{d}^{1:T})$ and marketing target $\mathbf{y}^{1:T}$.
To be specific, the joint distribution of causal structure is denoted as $\prod_{i=1}^{d+1} \prod_{j=1}^{d+1} z_{ij}$, where $z_{ij} = 1$ expresses that there exists a directed edge from node $i$ to $j$ (\ie, $e_{ij}$) and $i \neq j$. 
Since the underlying causal structures are not known in advance, we predict causal relations starting from a fully connected graph.
The graph neural network (GNN)~\cite{NRI/icml/KipfFWWZ18, NRI/iclr/Banijamali22} is utilized to propagate information across a fully connected graph and predict causal edges.
The causal relational encoder consists of three procedures, \ie, pairwise embedding, relational interaction, and Gumbel softmax sampling.

\subsubsection{Pairwise Embedding}

We first initialize the edge representation in the fully connected graph through pairwise embedding.
With a slight abuse of notation, $(\mathbf{X}, \mathbf{y})_j$ refers to record of the $j$-th channel $\mathbf{x}_j^{1:T}\ (1 \leq j \leq d) $, and the record of marketing target $\mathbf{y}^{1:T} \ (j = d+1)$ in the rest of the paper. And the formulation of pairwise embedding is as follows: 
\begin{align*}
    \mathbf{h}_{j}^1 &= f_{\mathrm{emb}}\big( (\mathbf{X}, \mathbf{y})_j  \big), \\
    \mathbf{h}_{ij}^1 &= f_{\mathrm{edge}}^1 \big( [\mathbf{h}_{i}^1, \mathbf{h}_{j}^1  ]   \big),
\end{align*}
where $\mathbf{h}_{j}^1$ denotes node representation and $\mathbf{h}_{ij}^1$ denotes pairwise representation.
$f_{\mathrm{emb}}$ and $f_{\mathrm{edge}}^1$ are fully-connected networks (MLPs).
It captures local information in a pairwise manner.

\subsubsection{Relational Interaction} To take the relational interaction with other nodes, \textit{a.k.a.}, global information, into account, we further calculate the edge embedding $\mathbf{h}_{ij}$ as formulated below: 
\begin{align*}
    \mathbf{h}_{j}^2 &= f_{\mathrm{vertex}}^1 \big( \sum_{i \neq j} \mathbf{h}_{ij}^1 \big), \\
    \mathbf{h}_{ij}^2 &= f_{\mathrm{edge}}^2 \big( [\mathbf{h}_{i}^2, \mathbf{h}_{j}^2 ] \big),
\end{align*}
where $f_{\mathrm{vertex}}^1$ and $f_{\mathrm{edge}}^2$ are implemented based on MLPs as well.     

\subsubsection{Gumbel Softmax Sampling} The above formulation can be summarized by $f_{\mathrm{enc}}(\cdot)$ to derive structure distribution as follows:
\begin{align*}
    \mathbf{h}_{ij} &= f_{\mathrm{enc}}( \mathbf{X}, \mathbf{y}   )  ,\\
           q_{\phi} \big(\mathbf{z} |  (\mathbf{X},\mathbf{y})  \big)  &= \mathrm{Softmax}( \mathbf{h}_{ij} / \tau )   ,
\end{align*}
where $\tau$ is the temperature parameter that controls the smoothness of sampling.  
As latent distribution $q_{\phi} \big(\mathbf{z} |  (\mathbf{X},\mathbf{y})  \big)$ are discrete and cannot be backpropagated through the reparametrization trick, we add Gumbel distributed noise~\cite{Gumbel/iclr/JangGP17} during training:
\begin{equation*}
    \mathbf{z}_{ij} \sim \mathrm{Softmax}\big(   (  \mathbf{h}_{ij} + \mathbf{g}  / \tau )    \big),
\end{equation*}
where $\mathbf{g}$ are i.i.d. samples drawn from a $\mathrm{Gumbel}(0,1)$ distribution.
$\mathbf{z}$ represents the inferred causal structures.

\subsection{Marketing Response Decoder}
\label{sec:M3}

\begin{figure}
    \centering

	\includegraphics[width=0.48\textwidth]{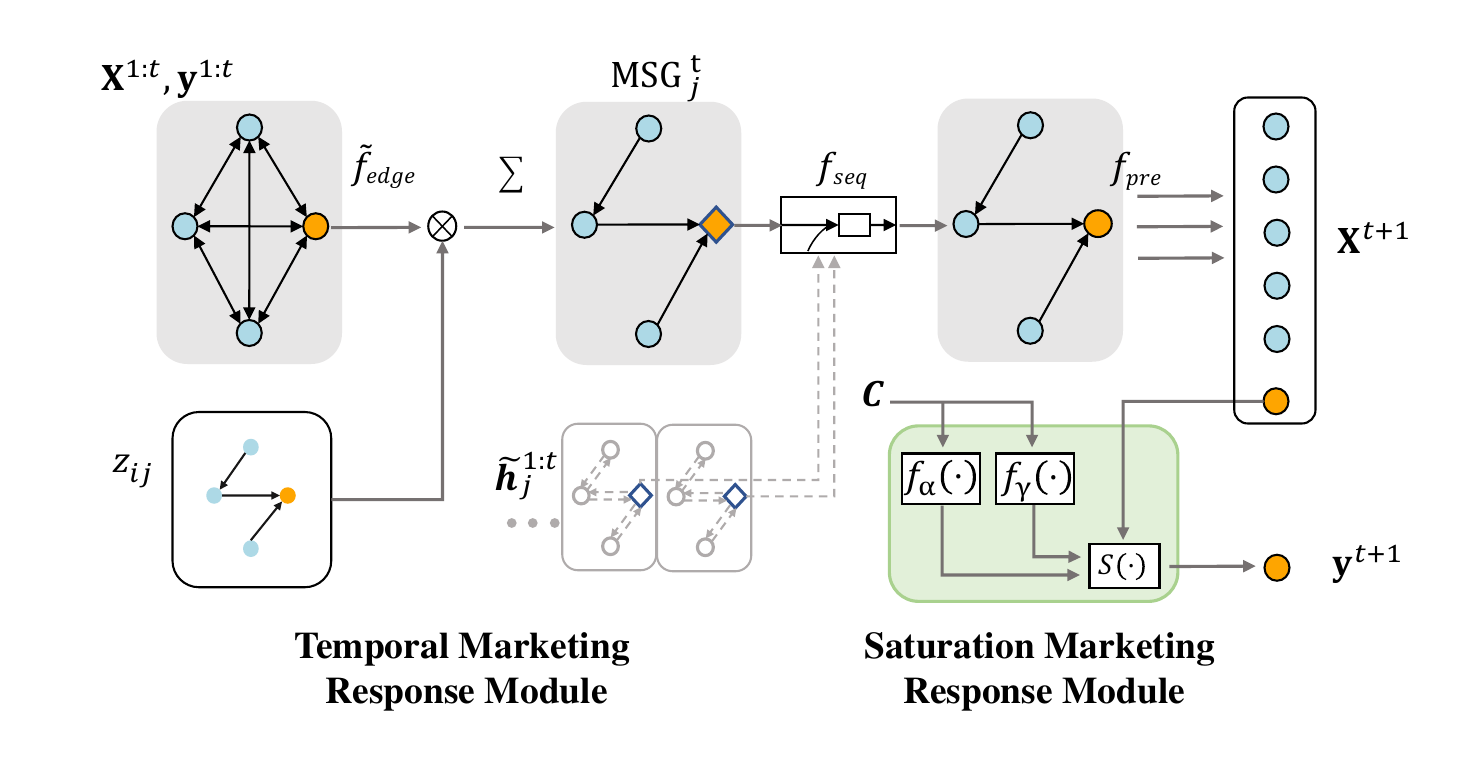} %
	\vspace{-6ex} 
	\caption{The structure of the marketing response decoder. At each step, the decoder takes the inferred causal structure $\mathbf{z}$, the past observation of $\mathbf{X,y}$, and the historical hidden states $\tilde{\mathbf{h}}_{}^{t+1}$ as input to model marketing response.}
	\vspace{-3ex}
         \label{fig:decoder}
\end{figure}

The aim of the decoder is to model marketing response under the inferred causal structures.
Although the complex causal interaction can be modeled owing to the above inferred causal structures, other marketing patterns are still unignorable.
There are two typical hypotheses in marketing mix modeling~\cite{MMM_46001/jin2017bayesian, comp_survey/wigren2019marketing}:

\hypothe{1}{Temporal Marketing Response}{Investments on advertising channels have lagged and decay effect over time, a.k.a, carryover effect.}

\hypothe{2}{Saturation Marketing Response}{Investments on advertising channels have diminishing returns.}

How to systematically model both temporal and saturation patterns is a nonnegligible issue.  
In most cases, channel cost at each time period is relatively small compared to that of cumulative channel cost across each time point. 
Therefore, we first model temporal patterns and then saturation patterns. 
As shown in Figure~\ref{fig:decoder}, the marketing response decoder consists of two procedures, \ie, temporal marketing response module, and saturation marketing response module.

\subsubsection{Temporal Marketing Response Module}

To incorporate temporal patterns, we add sequence models to the original message-passing mechanisms in GNNs. More formally:   
\begin{align}
    \label{eq_edgeSum}
    \tilde{\mathbf{h}}_{ij}^t &=  \tilde{f}_{\mathrm{edge}} \big([({\mathbf{X},\mathbf{y}})_{i}^t, ({\mathbf{X},\mathbf{y}})_{j}^t    ] \big),  \nonumber   \\
    \mathrm{MSG}_{j}^t  &=  \sum_{i \neq j}  z_{ij} \tilde{\mathbf{h}}_{ij}^t  ,  \\
    \tilde{\mathbf{h}}_{j}^{t+1} &= f_{\mathrm{seq}} \big(  \mathrm{MSG}_{j}^t,   \tilde{\mathbf{h}}_{j}^{1:t}  \big), \nonumber  \\ 
    \boldsymbol{\mu}_{j}^{t+1} &=  f_{\mathrm{pre}} \big(\tilde{\mathbf{h}}_{j}^{t+1}, (\mathbf{X},\mathbf{y})_j^t \big), \ \ j=1,...,d  \nonumber 
\end{align}
where the recurrent hidden states at the previous time step are leveraged for the message-passing mechanism.
$f_{\mathrm{seq}}$ is the sequence model implemented with RNN here, which takes $\mathrm{MSG}_{j}^t$ together with the current value $(\mathbf{X},\mathbf{y})_j^t$ and the previous hidden states $\tilde{\mathbf{h}}_{j}^{1:t}$ as input to capture the temporal pattern for each marketing variable. 
$f_{\mathrm{pre}}$ is modeled by MLPs.


\subsubsection{Saturation Marketing Response Module}
The diminishing return is vital to marketing decision-making as it reflects rational dynamics between marketing investment and response.
In this module, we focus on S-curve (Hill) transformation, which is the most widely used for modeling saturation\cite{MMM_46001/jin2017bayesian}, and we adopt it in a gradient-based way in predicting marketing target $\mathbf{y}^{t+1}$.

The vanilla S-curve $S(\cdot)$ for channels' saturation effect on $\mathbf{y}$ is defined as follows: 
\begin{equation}
\label{eq_scurve}
    \mathrm{Saturation\ effect} = \frac{ \mathrm{Temporal\ effect}^\alpha}{ \mathrm{Temporal\ effect}^\alpha + \gamma    }, \nonumber
\end{equation}
where $\alpha$ manipulates the shape of the curve between exponential and S-shape, whereas $\gamma$ indicates the inflexion point of the response curve.
Their values are influenced by the character of the market.  

We model $\alpha,\gamma$ through neural networks by taking the vector of contextual variable $\mathbf{C}$.  
This brings stronger representation power and interpretability to the model, as side information including shop type, event, and macro trend can be used to determine the shape of the saturation curve.
To be specific, the traditional S-curve model for the marketing target is extended to the following form: 
\begin{equation*}
    \boldsymbol{\mu}_{j}^{t+1} = \frac{ f_{\mathrm{pre}} \big(\tilde{\mathbf{h}}_{j}^{t+1}, (\mathbf{X},\mathbf{y})_{j}^t \big)^{f_\alpha( \mathbf{C} ) }     }{ f_{\mathrm{pre}} \big(\tilde{\mathbf{h}}_{j}^{t+1}, (\mathbf{X},\mathbf{y})_{j}^t \big)^{f_\alpha( \mathbf{C} ) } +  f_{\gamma}(\mathbf{C}) }, \ \ j=d+1
\end{equation*}
where $f_\gamma(\cdot)$ denotes a neural network to calculate the inflexion point of the market, and $f_\alpha(\cdot)$ is a neural network whose output controls the shape of the curve between exponential and S-shape. 
In this module, the capability of neural networks and the explicitness of S-curve is integrated.

Based on the above two modules, we can derive $\boldsymbol{\mu}_{1:d+1}^{t+1}$ characterizing the prediction value of $d$-variate channels and marketing target, respectively. 
The ultimate value is given as follows:
\begin{equation*}
    p_{\theta}\big( (\mathbf{X}^{t+1}, \mathbf{y}^{t+1})  | (\mathbf{X}^{1:t}, \mathbf{y}^{1:t}), \mathbf{z}  \big) =  \mathcal{N} \big( \boldsymbol{\mu}_{1:d+1}^{t+1}, \sigma^2 \mathbf{I}   \big),
\end{equation*}
where $\sigma$ is the fixed variance term.  
For multiple-step forecasting, the prediction results $\boldsymbol{\mu}_{j}^{t+m}$ is utilized in a recursive manner, \ie,
\begin{align*}
    \boldsymbol{\mu}^{t+1} &= f_{\mathrm{dec}}    \big(  (\mathbf{X},\mathbf{y})^t , f_{\mathrm{enc}}(\mathbf{X},\mathbf{y})  \big),   \\ 
    \boldsymbol{\mu}^{t+m} &= f_{\mathrm{dec}}    \big(   \boldsymbol{\mu}^{t+m-1} , f_{\mathrm{enc}}(\mathbf{X},\mathbf{y})  \big), \ \ \ \ m=2,...,M   
\end{align*}
where $M\ (M \geq 1)$ is the number of time steps to predict.

\subsection{Variational Inference for Optimization}
This section gives the optimization procedure with variational inference.
The parameters of $f_{\mathrm{enc}}$ and $f_{\mathrm{dec}}$ detailed above can be derived according to the following form :
\begin{equation*}
    f_{\mathrm{enc}*},f_{\mathrm{dec}*} = \mathrm{argmin}_{ f_{\mathrm{enc}},f_{\mathrm{dec}}} \mathcal{L}(\mathcal{D}_{\mathrm{}}, f_{\mathrm{enc}},f_{\mathrm{dec}}  ), 
\end{equation*}
where the loss function is composed of a data-fitting term and a structural regularization term, \ie,
\begin{align}  
    &\mathcal{L}(\mathcal{D}_{\mathrm{}}, f_{\mathrm{enc}},f_{\mathrm{dec}}  )  \nonumber \\
    =& \sum_n \sum_t {l} \Bigl( (  \mathbf{X}, \mathbf{y}  )^{t+1:t+M},   f_{\mathrm{dec}} \bigl(    (\mathbf{X}, \mathbf{y}  )^{1:t}, f_{\mathrm{enc}} \left( \mathbf{X}, \mathbf{y}   \right)   \bigr)   \Bigr) \nonumber  \\ 
    &+ \lambda \cdot  r \bigl( f_{\mathrm{enc}} \left( \mathbf{X}, \mathbf{y}   \right)  \bigr), \label{eq:loss_origin}
\end{align}
where $\lambda$ is a penalty factor of the structural prior. 
We further leverage variational inference to model the functions $f_{\mathrm{enc}}$ and $f_{\mathrm{dec}}$.
As the causal relational encoder $f_{\mathrm{enc}}$ via encoding function $q_{\phi} \big(\mathbf{z} |  (\mathbf{X},\mathbf{y})  \big)$ deriving a distribution over $\mathbf{z}$ which represents the predicted edges in the causal structure, and a decoder $p_{\theta}\big(  ( \mathbf{X}, \mathbf{y}  )     | \mathbf{z}  \big)$ that probabilistically models marketing response under the inferred causal structure. 
Thus, the loss function $\mathcal{L}(\mathcal{D}_{\mathrm{}}, f_{\mathrm{enc}},f_{\mathrm{dec}})$ in Equation~\ref{eq:loss_origin} can be reformulated as a variational lower bound:
\begin{equation}
        \mathcal{L} = 
    \mathbb{E}_{ q_{\phi} (\mathbf{z} |  (\mathbf{X},\mathbf{y})  )   } \big[ \mathrm{log}   p_{\theta}\big(  ( \mathbf{X}, \mathbf{y}  )     | \mathbf{z}  \big)     \big]
    - \lambda \cdot \mathrm{KL} \big[ q_{\phi} \big(\mathbf{z} |  (\mathbf{X},\mathbf{y})  \big)   || p(\mathbf{z})      \big]    , \nonumber
\end{equation}
where the first term is a negative log-likelihood for data fitting, and the second term is a KL divergence to a structural prior distribution for regularization.


\subsection{Theoretical Analysis of \our}
\label{sec:M4} 
This section first gives the theoretical analysis that we can infer Granger causality in the framework of \our. Then the computational complexity of the proposed method is analyzed.

Aligning with the definition of non-linear Granger causality, our \our is on the premise that there exists some function $g$ describing the marketing response of any shop or brand $\forall n, (1\leq n \leq N)$ given its historical marketing mix dataset $(\mathbf{X}^{}_n, \mathbf{y}^{}_n)^{1:t}$ and its underlying causal structure $\mathcal{G}_n$: $(\mathbf{X}^{}, \mathbf{y}^{}  )^{t+1}  = g \big(  (\mathbf{X}^{}, \mathbf{y}^{}  )^{1:t}, \mathcal{G}_n   \big) + \boldsymbol{\epsilon}^{t+1}_n$.
The unknown components of the data-generation process above can be divided into two parts: (1) the causal structure $\mathcal{G}_n$ specific to the $n$-th shop; (2) the mapping $g$ of marketing response.
We make the following claim, and details of our proof can be found in Appendix~\ref{app:GC_proof}.


\claim{1}{Granger Causality of \our}{For the $n$-th sample in dataset, the variable $\mathbf{x}_{n,i}$ does not Granger cause variable $\mathbf{x}_{n,j}$, if $z_{n,ij}=0$ according to \our.}

We then discuss the computational complexity of the proposed \our in both training and causal relations inference stages. For a shop's marketing mix records consisting of $T$ days, their time complexities are $\mathcal{O}(WT)$ and $\mathcal{O}(W)$, respectively, where $W$ is the number of weights.
A detailed analysis of each component can be found in Appendix~\ref{app:complexity_ana}.
The time complexity of the whole model is $\mathcal{O}(WT)$.  
Appendix~\ref{app:running_time} provides a comparison between training time on simulated datasets with varying sizes. According to the results, \our scales linearity with the increase in training data size. Thus, it's promising to handle massive marketing data from a large number of shops.

%% file: sections/experiment.tex
\section{Experiment}

In this section, we evaluate the performance of \our and answer the following research question:  
\begin{itemize}
    \item \textbf{RQ1:} As a framework based on temporal causal discovery, can \our accurately recover causal structures from heterogeneous marketing data?  
    \item \textbf{RQ2:} What is the performance of \our in terms of predicting the target variable?     
    \item \textbf{RQ3:} Can causal structures discovered by \our align with expert knowledge on real-world datasets?      
    \item \textbf{RQ4:} What are the capabilities of the causal relational encoder, the temporal marketing response module, and the saturation marketing response module?  
\end{itemize}

\subsection{Experimental Settings}
This section provides an overview of the data, experimental protocol, evaluation metrics, and compared baselines.

\begin{table}[h]
    \centering
    \caption {Dataset statistics}
    \label{tab:dataset}
    \vspace{-2ex}
    \small
 \begin{tabular}{c|cccc}
  \toprule  
    Dataset & Shops & Channels & Context variables & Length \\
    \midrule
    Simulation 1 & 50 & (5,10,20) & - & (30,120,720) \\
    Simulation 2 & 100 & (5,10,20) & 2 & (30,120,720) \\
    AirMMM & 50 & 12 & 5 & 676 \\
  \bottomrule  
 \end{tabular}
 \vspace{-4ex}  
\end{table}

\subsubsection{Data Descriptions.} 
The performance of \our is evaluated on two datasets.
The first dataset is a synthetic dataset created to test \our's ability to reconstruct causal structures. 
The data generation procedure consists of three steps, \ie, graph sampling, channel generation, and response generation. 
Compared to \textbf{Simulation 1}, the saturation curve in \textbf{Simulation 2} is characterized by contextual variables.
We further adjust the channel numbers and lengths in simulation to test the effectiveness of \our under different conditions.
The detailed data generation procedure can be found in Appendix~\ref{app:sim_data}. 
The second, \textbf{AirMMM}, is a real-world marketing mix dataset collected from an E-commerce platform, which contains 50 shops' marketing mix data with a period of $22$ months from Jan 30th 2021 to Dec 6th 2022.
Appendix~\ref{app:real_data} describes the details of advertising channels in AirMMM. 
The statistics on two marketing mix datasets are shown in Table~\ref{tab:dataset}.

\subsubsection{Experimental Protocol.} We conduct experiments on two data- sets to evaluate \our's performance. 
For the synthetic dataset, we simulate heterogeneous longitudinal data in settings with and without saturation patterns and we quantify the ability of \our to reconstruct causal structures (Section~\ref{exp_sec:RQ1}).
For the real-world dataset, we compare \our's performance in terms of GMV prediction to that of two categories of state-of-the-art methods, \ie, marketing mix models and temporal causal discovery models (Section~\ref{exp_sec:RQ2}).
We also report the inferred causal structures on real-world datasets, which align with domain experts' knowledge (Section~\ref{exp_sec:RQ3}). 
Moreover, we provide the ablation and parameter sensitivity studies for verifying the effectiveness and robustness of each component (Section~\ref{exp_sec:RQ4}).

\subsubsection{Evaluation Metrics.}
For evaluating the learned causal structure against ground truth, two metrics are reported.
Accuracy (\textbf{ACC}), and Areas Under Receiver Operating Characteristic (\textbf{AUROC}). 
AUROC is defined as the ratio of true and false positive rates given the threshold varies between $0$ and $1$. 
The prediction of self-connectivity (\ie, the diagonal part of the adjacency matrix) is not evaluated in our experiment, which is the easiest relation to infer.
The reported results are averaged over ten random trials.
For evaluating the performance of GMV forecasting, the Mean Squared Error (\textbf{MSE}) metric is used. 

\subsubsection{Compared Methods}

To answer the first question (RQ1), we compare our \our method with the following popular baselines:
\begin{itemize}
    \item \textbf{Linear Granger}~\cite{kdd/ArnoldLA07}. As one of the most well-known methods, it applies the vector autoregressive (VAR) model with ridge regularization to learn causal structures.
    \item \textbf{NGC}~\cite{MTS/Granger/pamiNGC22}. It leverages an LSTM or MLP to predict the future and conduct causal discovery based on input weights. 
    \item \textbf{GVAR}~\cite{MTS/Granger/iclr21_GVAR_MarcinkevicsV}. It's an interpretable Granger causal structure learning method which considers both sign effect and time reversal. Causal links are derived based on self-explaining neural networks. 
    \item \textbf{InGRA}~\cite{CD/icdm/InGRA_ChuWMJZY20}. It aims to learn Granger causal structures from heterogeneous MTS based on attention mechanism and prototype learning. 
\end{itemize}

To answer the second question (RQ2), we additionally compare \our in terms of GMV forecasting with several competitive methods as follows:
\begin{itemize}
    \item \textbf{LSTM}~\cite{forLSTM/neco/HochreiterS97}. Its ability to process sequential data with varying time lags can be utilized for GMV forecasting.
    \item \textbf{Wide \& Deep}~\cite{forWideDeep/recsys/Cheng0HSCAACCIA16}. It combines regression and DNN that benefits from interpretability and representation power.
    \item \textbf{BTVC}~\cite{adkdd/ng2021bayesian}. It's a time-varying coefficient model based on hierarchical Bayesian structures for MMM. 
\end{itemize}

To verify the effectiveness of causal relational encoder, temporal marketing response module, and saturation marketing response module (RQ4), we introduce some variants of \our as follows:
\begin{itemize}
    \item \textbf{\ourfu}. It replaces the causal relational encoder with a full graph input to the decoder. 
    \item \textbf{\ourmar}. It replaces the temporal marketing response module with a Markovian module that only captures the records of the last time step.
    \item \textbf{\ourrw}. It's a variant of our method without the saturation marketing response module.
\end{itemize}

\subsection{Performance of Causal Structure Learning (RQ1)} \label{exp_sec:RQ1}

\begin{table*}[t]
\centering
\setlength\abovecaptionskip{0pt}
\caption{Causal structure reconstruction results with different simulation settings.} 
\label{tab:RQ1Res1}
\begin{tabular}{ccccc}
\toprule
\multirow{2}{*}{Methods} & \multicolumn{2}{c}{Simulation 1}                                 & \multicolumn{2}{c}{Simulation 2}                                \\ \cline{2-5} 
                         & ACC                       & AUROC                      & ACC                       & AUROC                      \\ \hline
Linear Granger                 & 0.630$\pm$0.101 & 0.628$\pm$0.147 & 0.624$\pm$0.196 & 0.661$\pm$0.020 \\

NGC                 & 0.667$\pm$0.251 & 0.694$\pm$0.170 & 0.596$\pm$0.235 & 0.639$\pm$0.265 \\

GVAR            & 0.839$\pm$0.110 & 0.847$\pm$0.092 & 0.784$\pm$0.148 & 0.859$\pm$0.149 \\

InGRA      & {0.877$\pm$0.096} & 0.863$\pm$0.082 & 0.856$\pm$0.098 & 0.854$\pm$0.083 \\   

\ourrw      & 0.921$\pm$0.037 & 0.924$\pm$0.049 & 0.858$\pm$0.053 & 0.871$\pm$0.071 \\

\ourmar      & 0.872$\pm$0.039 & 0.887$\pm$0.051 & 0.832$\pm$0.087 & 0.865$\pm$0.062 \\

\our                   & \textbf{0.923$\pm$0.020} & \textbf{0.935$\pm$0.012} &   \textbf{0.892$\pm$0.033} & \textbf{0.903$\pm$0.015} \\

\bottomrule
\end{tabular}
\vspace{-1ex}
\end{table*}

\begin{table*}[t]
\centering\setlength\abovecaptionskip{0pt}
\caption{Causal structure reconstruction results w.r.t the channel number ($R$=5, $T$=120).}
\label{tab:FixT_VarNode}
\begin{tabular}{ccccccccc}
\toprule
\multirow{2}{*}{Methods} & \multicolumn{2}{c}{\textit{d}=5} & \multicolumn{2}{c}{\textit{d}=10} & \multicolumn{2}{c}{\textit{d}=20}  \\ \cline{2-7} 
                             & ACC     & AUROC     & ACC     & AUROC     & ACC     & AUROC     \\ \hline

Linear Granger                     &    0.612$\pm$0.125   &    0.644$\pm$0.017    &     0.624$\pm$0.018       &    0.661$\pm$0.020         &    0.580$\pm$0.116        &        0.603$\pm$0.058     \\
               
NGC                              &    0.623$\pm$0.278       &      0.665$\pm$0.151     &         0.596$\pm$0.235   &     0.639$\pm$0.265        &           0.534$\pm$0.212     &  0.595$\pm$0.120            \\

GVAR                             &    0.842$\pm$0.158       &      0.884$\pm$0.072     &         0.784$\pm$0.148   &     0.859$\pm$0.149        &           0.792$\pm$0.146     &  0.878$\pm$0.084            \\

InGRA                             &    0.897$\pm$0.181       &      {0.882$\pm$0.140}     &         0.856$\pm$0.098   &     0.854$\pm$0.083        &           0.839$\pm$0.024     &  0.872$\pm$0.059            \\  

\ourrw                             &   0.885$\pm$0.178	    &  {0.842$\pm$0.153}      &    	{0.858$\pm$0.053}       &  	{0.871$\pm$0.071}	         &       {0.836$\pm$0.025}     &        	{0.860$\pm$0.045}        \\

\ourmar      & 0.854$\pm$0.157 & 0.839$\pm$0.197 & 0.832$\pm$0.087 & 0.865$\pm$0.062 & 0.814$\pm$0.027 & 0.825$\pm$0.032 \\

\our       	            &   \textbf{0.925$\pm$0.140}    &  \textbf{0.910$\pm$0.187}      &   	\textbf{0.892$\pm$0.033}	         &    \textbf{0.903$\pm$0.015}	         &   \textbf{0.875$\pm$0.109}	         &     \textbf{0.905$\pm$0.024}        \\

\bottomrule
\end{tabular}
\vspace{-1ex}
\centering
\end{table*}





\begin{table*}[t]
\centering
\setlength\abovecaptionskip{0pt}\setlength{\abovecaptionskip}{0pt} 
\setlength{\belowcaptionskip}{0pt} 
\caption{Causal structure reconstruction results w.r.t the series length  ($R$=5, $d$=10). }
\label{tab:FixNode_VarT}
\begin{tabular}{@{}ccccccccc@{}}
\toprule
\multirow{2}{*}{Methods} & \multicolumn{2}{c}{\textit{T}=30} & \multicolumn{2}{c}{\textit{T}=120} & \multicolumn{2}{c}{\textit{T}=720}  \\ \cline{2-7} 
                             & ACC     & AUROC     & ACC     & AUROC     & ACC     & AUROC     \\ \hline

Linear Granger                     &  0.514$\pm$0.000     &   0.581$\pm$0.000    &           0.624$\pm$0.196     &   0.661$\pm$0.020            &          0.735$\pm$0.152       &    0.892$\pm$0.180             \\

NGC                          &     0.573$\pm$0.051     &    0.568$\pm$0.074        &  0.596$\pm$0.235        &    0.639$\pm$0.265       &            0.744$\pm$0.149        &      0.752$\pm$0.185            \\

GVAR                             &    0.693$\pm$0.116       &      0.716$\pm$0.095     &         0.784$\pm$0.148   &     0.859$\pm$0.149        &           0.886$\pm$0.084     &  0.890$\pm$0.072            \\

InGRA                              &    0.774$\pm$0.158       &      0.763$\pm$0.142     &         0.856$\pm$0.098   &     0.854$\pm$0.083        &           0.899$\pm$0.075     &  0.907$\pm$0.061            \\ 

\ourrw                             &   0.800$\pm$0.110	    &  {0.815$\pm$0.105}      &    	{0.858$\pm$0.053}       &  	{0.871$\pm$0.071}	         &       {0.914$\pm$0.008}     &        	{0.932$\pm$0.015}        \\

\ourmar      & 0.763$\pm$0.103 & 0.787$\pm$0.085 & 0.832$\pm$0.087 & 0.865$\pm$0.062 & 0.868$\pm$0.023 & 0.879$\pm$0.019 \\

\our       	            &   \textbf{0.843$\pm$0.119}    &  \textbf{0.852$\pm$0.127}      &   	\textbf{0.892$\pm$0.033}	         &    \textbf{0.903$\pm$0.015}	         &   \textbf{0.941$\pm$0.031}	         &     \textbf{0.952$\pm$0.026}        \\

\bottomrule
\end{tabular}
\vspace{-1ex}
\end{table*}

\begin{table*}[t]
\centering\setlength\abovecaptionskip{0pt}
\caption{Causal structure reconstruction results w.r.t. the latent structure number $R$ ($d$=10, $T$=120).}
\label{tab:VARstructureNo}
\begin{tabular}{@{}ccccccc@{}}
\toprule
\multirow{2}{*}{Methods}                                      & \multicolumn{2}{c}{\textit{R}=5}                                     & \multicolumn{2}{c}{\textit{R}=10}                                     & \multicolumn{2}{c}{\textit{R}=20}                                    \\ \cline{2-7} 
                                               & ACC     & AUROC     & ACC     & AUROC     & ACC     & AUROC      \\ \hline
InGRA                  & 0.876$\pm$0.098 & 0.874$\pm$0.083 & 0.848$\pm$0.180 & 0.865$\pm$0.119 & 0.815$\pm$0.137 & 0.820$\pm$0.158 \\
\our           & \textbf{0.892$\pm$0.033} & \textbf{0.903$\pm$0.015} & \textbf{0.918$\pm$0.057} & \textbf{0.925$\pm$0.104} &  \textbf{0.877$\pm$0.066}  & \textbf{0.882$\pm$0.029} \\ \bottomrule
\end{tabular}

\end{table*}

\begin{figure}[t]
    \centering
	\includegraphics[width=0.36\textwidth]{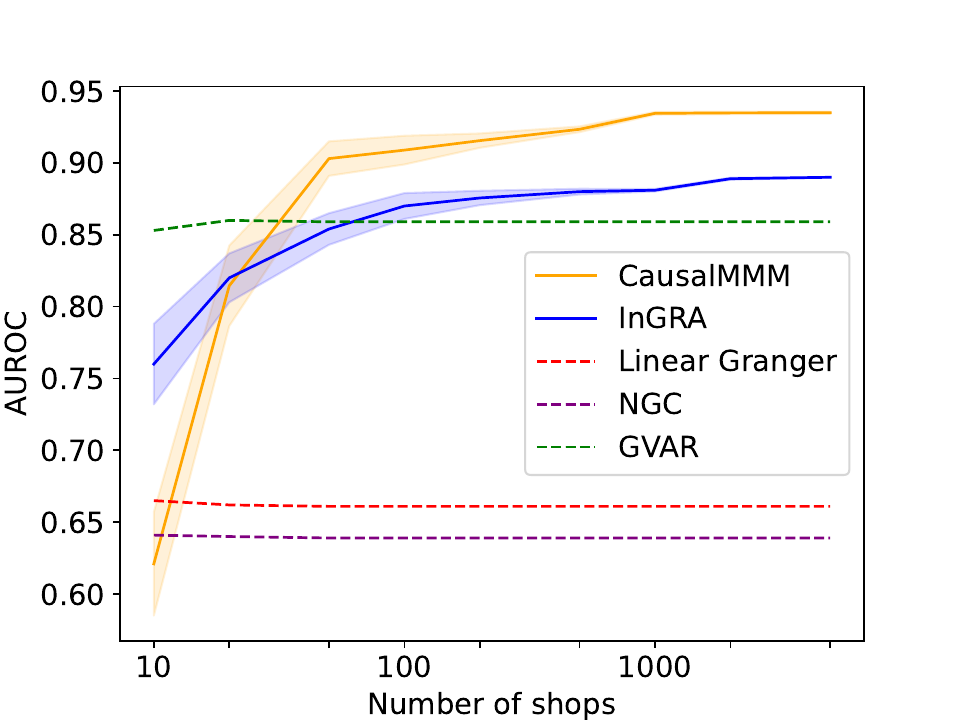} %
	\caption{Causal structure learning performance (in AUROC) \textit{w.r.t.} the number of shops $N$ ($d=10, T=120$).}
	\vspace{-2ex}
	\label{fig:Scalability}
\end{figure}

To answer RQ1, we conduct the synthetic experiment.
The main results of causal structure learning are shown in Table~\ref{tab:RQ1Res1}. 
To further exploit the effectiveness of our \our under heterogeneous data, we also report ACC and AUROC results in simulation 2 \textit{w.r.t.} the channel number $d$, the series length $T$, and the number of latent causal structures $R$ in Table~\ref{tab:FixT_VarNode} to~\ref{tab:VARstructureNo}. 
In Figure~\ref{fig:Scalability}, we illustrate the model's performance with the increasing number of shops.
 
As shown in Table~\ref{tab:RQ1Res1}, \our consistently outperforms the compared baselines with AUROC improvements of $5.7\%\sim7.1\%$, which demonstrates its capability in causal structure learning. 
InGRA is the strongest baseline but also inferior to \our. 
It leverages prototype learning, which is helpful to extract prototypical structures in heterogeneous data.  
However, InGRA does not consider the patterns of marketing response, which would harm the performance, especially in Simulation 2 where a more complicated marketing response is entailed.
In Table~\ref{tab:FixT_VarNode}, we vary $d$ and generate heterogeneous datasets. 
As shown, \our consistently outperforms compared methods across all cases and achieves good performance even when $d$ reaches $20$.
It demonstrates that \our effectively handles complex structures with a large number of nodes.
In real scenarios, data collected from different shops may have varying lengths. 
Thus, we also vary the lengths in the evaluation. 
As shown in Table~\ref{tab:FixNode_VarT}, our \our still outperforms all compared baselines across different series lengths.
We also observe that the decreases in $T=30$, but \our exhibits a relatively smaller degree of degradation compared to baselines such as NGC and GVAR.
In Table~\ref{tab:VARstructureNo}, by adjusting $R$ we directly control the heterogeneity. 
We witness a performance decrease for InGRA with increasing $R$, which is due to the constraints of a pre-defined prototype number.
\our outperforms InGRA with the optimal performance achieved at $R=10$.
It demonstrates \our's capability in causal structure learning on datasets with more heterogeneous structures.

\begin{figure}
    \centering
    \vspace{-1ex} 
	\includegraphics[width=0.32\textwidth]{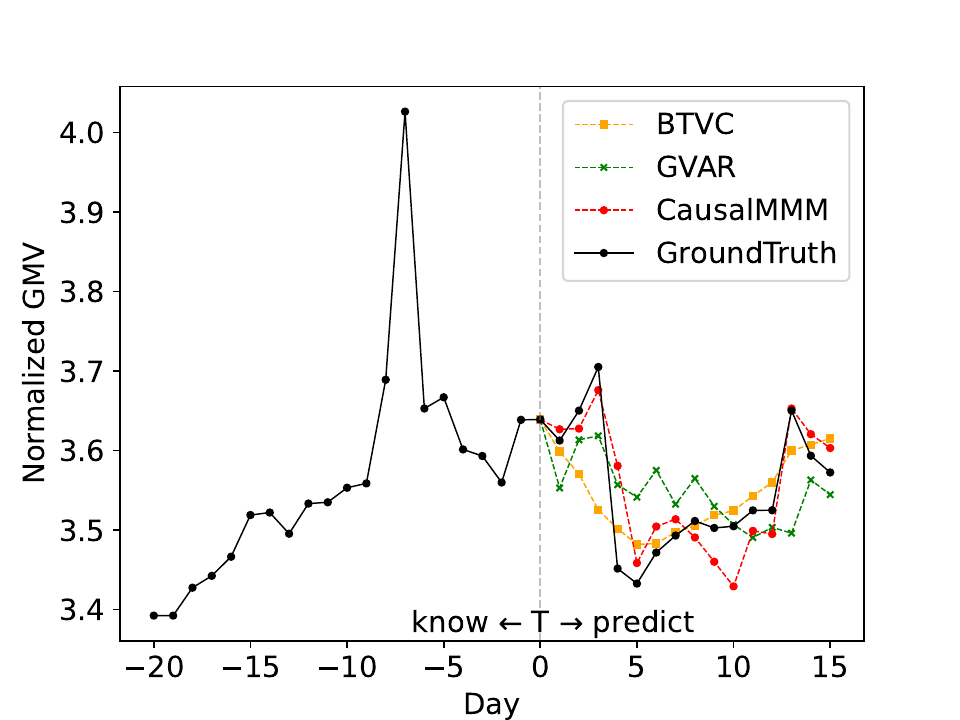} %
	\vspace{-2ex} 
	\caption{Visualization results for GMV prediction.}
	\vspace{-4ex}
	\label{fig:real_Viz_GMV_for}
\end{figure}

To explore the model's ability to utilize heterogeneous samples, the performance is illustrated in Figure~\ref{fig:Scalability} with the increase in $N$. 
Although \our is inferior to baselines such as GVAR and InGRA in low-data settings, its performance improves steadily when $N$ increases.
It significantly outperforms other baselines when $N \geq 100$.
This also demonstrates \our's capability in modeling heterogeneous data.
And it's meaningful in real-world scenarios where the marketing mix data across various shops are leveraged for sales promotion.

\subsection{Performance of GMV Prediction (RQ2)} \label{exp_sec:RQ2}
To answer RQ2, we employ a real-world dataset AirMMM to evaluate the performance of \our.
A detailed comparison and a visualization of the predicted results will be presented, respectively.

We show MSE results in Table~\ref{tab:pred_mse}, in which methods are compared under different prediction steps, \ie, $1,7,30$.   
We can observe that: 
(1) Among all the compared methods, BTVC performs best at $M=30$, which could be attributed to the helpful role of explicitly incorporating trend and seasonality information. 
(2) Our \our achieves the best performance at $M=7$, while attaining the second-best performance at $M=1,30$, which demonstrates the effectiveness of the proposed method. 

We also provide the visualization results in Figure~\ref{fig:real_Viz_GMV_for}, including the future $15$-days forecasts of the BTVC, GVAR, and the proposed \our.
As we can see, when predicting the future $15$-days, the \our can catch the sudden alteration and perform good forecasting in multi-steps with the help of both the marketing response pattern characterized by contextual variables and learned causal structures.

\begin{table}[t!]
\centering
\caption{GMV prediction results. The `-' denotes that multi-step forecasting is not supported in original implementation.}  
\vspace{-2ex}
\begin{tabular}{@{}cccc@{}}
\toprule
\multirow{2}{*}{Methods}  & \multicolumn{3}{c}{MSE} \\ \cline{2-4} 
&  $M=1$ & $M=7$ & $M=30$ \\ \hline

Linear Granger   & 0.60 & 5.96 & 21.65  \\
NGC   & 0.55 & - & -  \\
GVAR   & 0.37 & \underline{1.91} & 12.41 \\
InGRA   & 0.48 & - & -\\    
\hline
LSTM   & 0.58 & 3.31 & 15.78  \\
Wide \& Deep   & \textbf{0.25} & - & - \\
BTVC   & 0.47 & 2.07 & \textbf{9.43}  \\   
\hline

\ourfu   & 0.37 & 2.65 & 14.63  \\
\ourmar   & 0.32 & 2.85 & 12.92  \\ 
\ourrw   & 0.38 & 3.09 & 12.74  \\ 
\our   & \underline{0.29} & \textbf{1.80} & \underline{9.55}  \\
\bottomrule
\end{tabular}
\vspace{-2ex}
\label{tab:pred_mse}
\end{table}

\subsection{Graph Visualization (RQ3)} \label{exp_sec:RQ3}

\begin{figure}[h]   
    \centering
	\includegraphics[width=0.41\textwidth]{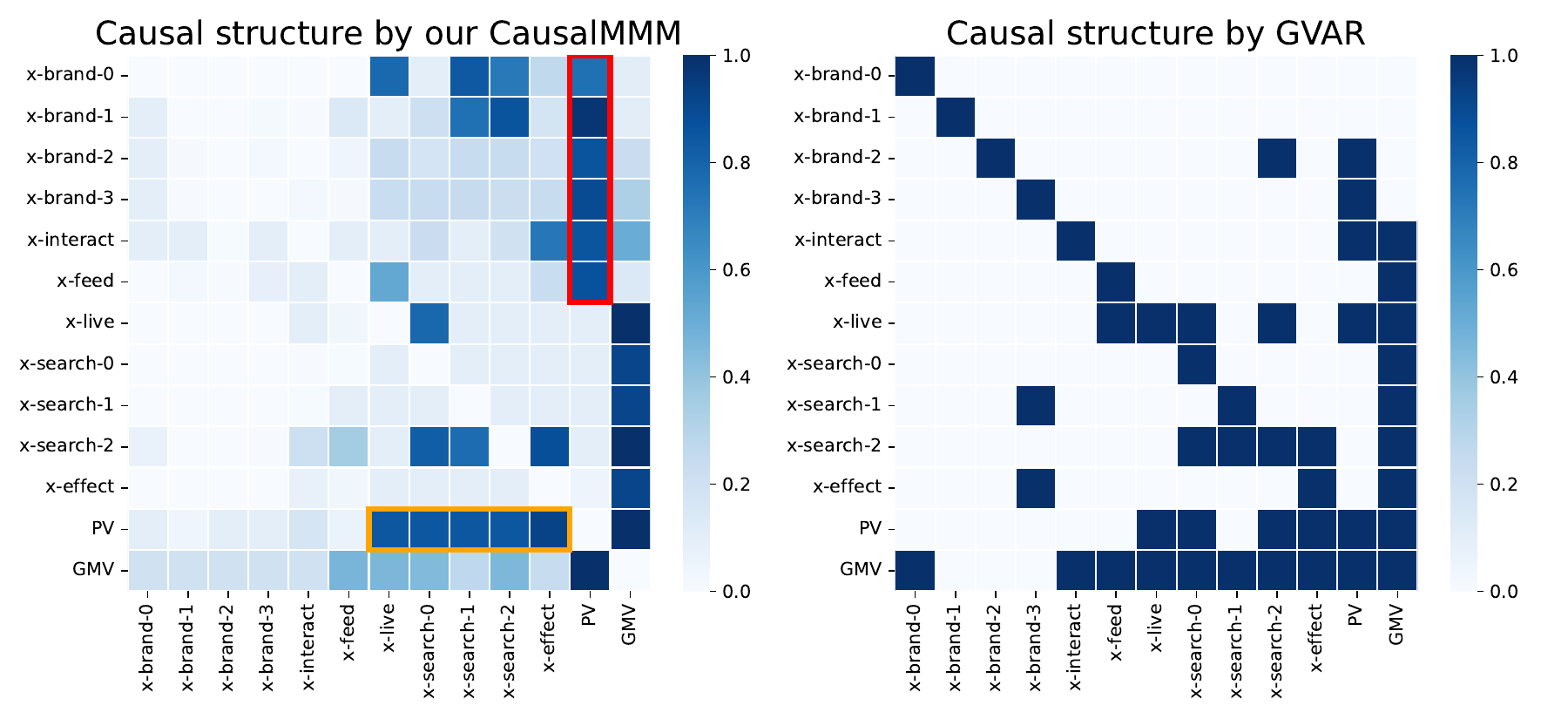} %
	\vspace{-4ex} 
	\caption{Learned causal structure of AirMMM. Causal relations from brand channels to PV are highlighted in {\red{red}}, and those from PV to effect channels are highlighted in {\color{orange}orange}.}
	\vspace{-6.5ex}
	\label{fig:Graph_onRealData_Viz}
\end{figure}

To answer RQ3, this section visualizes the learned causal structure of a beauty store on real-world data.
We illustrate the causal structures of $11$ channels, page view (PV), and the marketing target (\ie, GMV) by \our and the best from compared methods in Figure~\ref{fig:Graph_onRealData_Viz}.
Among the channels, the first six (\ie, x-brand-0, ..., x-feed) are brand channels designed to promote users' awareness and interest. In contrast, the remaining five (\ie, x-live, ..., x-effect) are effect channels where users are more likely to take action and convert. 
The details of these channels, including the media type and the means of placement, are described in Appendix~\ref{app:real_data}.


As we can see, causal relations from brand channels to PV, and PV to effect channels are observed.
These results agree with expert knowledge and are consistent with the conclusions of the marketing funnel effect~\cite{MMM_google/chan2017challenges}.
Our method also suggests several meaningful relations, including x-brand-1 to x-search-1/2, which indicate the increase in search volume on the E-commerce platform due to ads placements on the other video website. 
Such causal relations, however, can never be detected or utilized in traditional MMM methods.
Compared against \our, GVAR and the other baselines fail to discover these interactions.



\subsection{Ablation and Parameter Study (RQ4)} \label{exp_sec:RQ4}
To answer RQ4, we first compare \our with three ablation methods. 
The parameter-sensitive analysis is also included.

We summarize the metrics of ablation studies in Table~\ref{tab:RQ1Res1} and Table~\ref{tab:pred_mse}.
As shown, the AUROCs of \ourrw, \ourmar are inferior to \our, which proves that both the saturation marketing response module and the temporal marketing response module help improve causal structure learning.
We can observe that the performance of \ourrw is much better than that of \ourmar, which indicates the temporal pattern is more informative compared to the saturation pattern in causal structure learning.
For GMV forecasting, \ourfu, \ourmar, and \ourrw have larger prediction errors than \our, which demonstrates the effectiveness the causal relational encoder, the temporal marketing response module, and the saturation response module.
We also observe the improvement of the causal relational encoder is more significant than the other two modules of the decoder when $M=30$, which indicates that structural information is important for long-term prediction.  

The hyperparameter $\lambda$ is critical to balance data fitting and structural regularization. 
The temperature factor $\tau$ controls the ``smoothness" of Gumbel softmax sampling.
We conduct extensive experiments across a wider range of $\lambda$ and $\tau$, and the results are shown in Figure~\ref{fig:ParaStudy}. 
AUROC increases steadily and peaks around $\lambda = 10^2$.   
It verifies that adjusting $\lambda$ can control the effect of structural prior and reasonable values will benefit causal structure learning.
Its performance is relatively stable when the softmax temperature $\tau$ is not too small, which avoids degenerating to one-hot sampling.


%% file: sections/related.tex
\section{Related Work} 
\subsection{Marketing Mix Modeling}
Marketing mix modeling leverages statistical techniques and historical data to predict marketing targets with respect to advertising investment~\cite{MMM_google/chan2017challenges, comp_survey/wigren2019marketing}. 
According to whether assuming the causal structures, existing methods can be divided into two categories.
Approaches in the first category~\cite{comp_survey/wigren2019marketing, MMM_google/wang2017hierarchical} utilize linear regression or its non-linear variants, on the premise that regressors are mutually independent. 
Recent years have witnessed a vast amount of methods~\cite{MMM_interact/vaver2017introduction, MMM_46001/jin2017bayesian, adkdd/ng2021bayesian, JAS/chen2021hierarchical, MMM_google/sun2017geo} in this category. 
Albeit extant work~\cite{MMM_google/chan2017challenges, adkdd/ng2021bayesian} raises the multicollinearity concern, it neglects the complex causal relations among marketing variables.
Another category assumes a predefined causal model~\cite{MMM_google/chen2018bias} that aligns with the channels' funnel effects.
Given causal diagrams for the paid search channel, causal effects of advertising on sales are estimated with bias correction in~\cite{MMM_google/chen2018bias}.
However, in real-world scenarios with various advertising channels, the latent causal structures may be complex and heterogeneous across shops.
And no extant work tackles this issue.

\subsection{Causal Discovery from Temporal Data}

Causal reasoning~\cite{pearl2009causality} is an incredibly valuable tool that finds extensive application in the field of advertising and online marketing~\cite{NN_bg/kdd/YaoGZCB22, DBLP:conf/wsdm/ChengG022, DBLP:conf/wsdm/CaiFWYLX23, DBLP:conf/wsdm/TanZHKYYZYPZ022}.
The idea of enhancing MMM by modeling causality among marketing variables is inspired by works in causal discovery from temporal data~\cite{intro/ts_surveys/AssaadDG22, intro/ts_surveys/Moraffah21, KDD_tutorial_ours, csur/VowelsCB23, nips/WeichwaldJMPTV19}. 
Existing works can be grouped into four categories, \ie, constraint-based methods~\cite{MTS/CB/LPCMCI_GerhardusR20, MTS/CB_FCI_SVAR_FCI_MalinskyS18}, score-based methods~\cite{MTS/SB/NTS_NOTEARS, MTS/SB/Dynotears_aistats_PamfilSDPGBA20}, functional causal model (FCM)-based methods~\cite{MTS/FCM/cikm_WuWWLC22, MTS/FCM/nips_PetersJS13}, and Granger causality methods~\cite{MTS/Granger/pamiNGC22, MTS/Granger/iclr21_GVAR_MarcinkevicsV}.
Among them, Granger causality~\cite{granger1969investigating, Granger_review} is a popular and practical tool for causal analysis in many real-world applications~\cite{NN_bg_causal/www/NuaraSTZ0R19, DBLP:conf/cikm/ArabzadehFZNB18}.
To extract non-linear relations in high-dimensional conditions, a series of works~\cite{MTS/Granger/pamiNGC22,  MTS/Granger/iclr21_GVAR_MarcinkevicsV, MPIR/tailin, Discussion/NewForm/ACD_LoweMSW22, CD/icdm/InGRA_ChuWMJZY20} have been proposed recently, including methods based on information regularization~\cite{MPIR/tailin}, component-wise modeling~\cite{MTS/Granger/pamiNGC22, MTS/Granger/iclr20_esru}, low-rank approximation~\cite{MTS/Granger/SCGL_CIKM_XuHY19}, self-explaining networks~\cite{MTS/Granger/iclr21_GVAR_MarcinkevicsV}, and inductive modeling~\cite{CD/icdm/InGRA_ChuWMJZY20}.
Most extant works train a separate model for each sample which cannot take advantage of the information shared in the whole dataset, except for InGRA which utilizes prototype learning in heterogeneous time series.
Besides, most extant work only designed for causal structure learning neglects domain-specific patterns in real-world applications.

Different from existing studies, \our can discover heterogeneous causal structures while complying with the prior marketing response patterns. 


\begin{figure}[t] 
    \centering
	\includegraphics[width=0.48\textwidth]{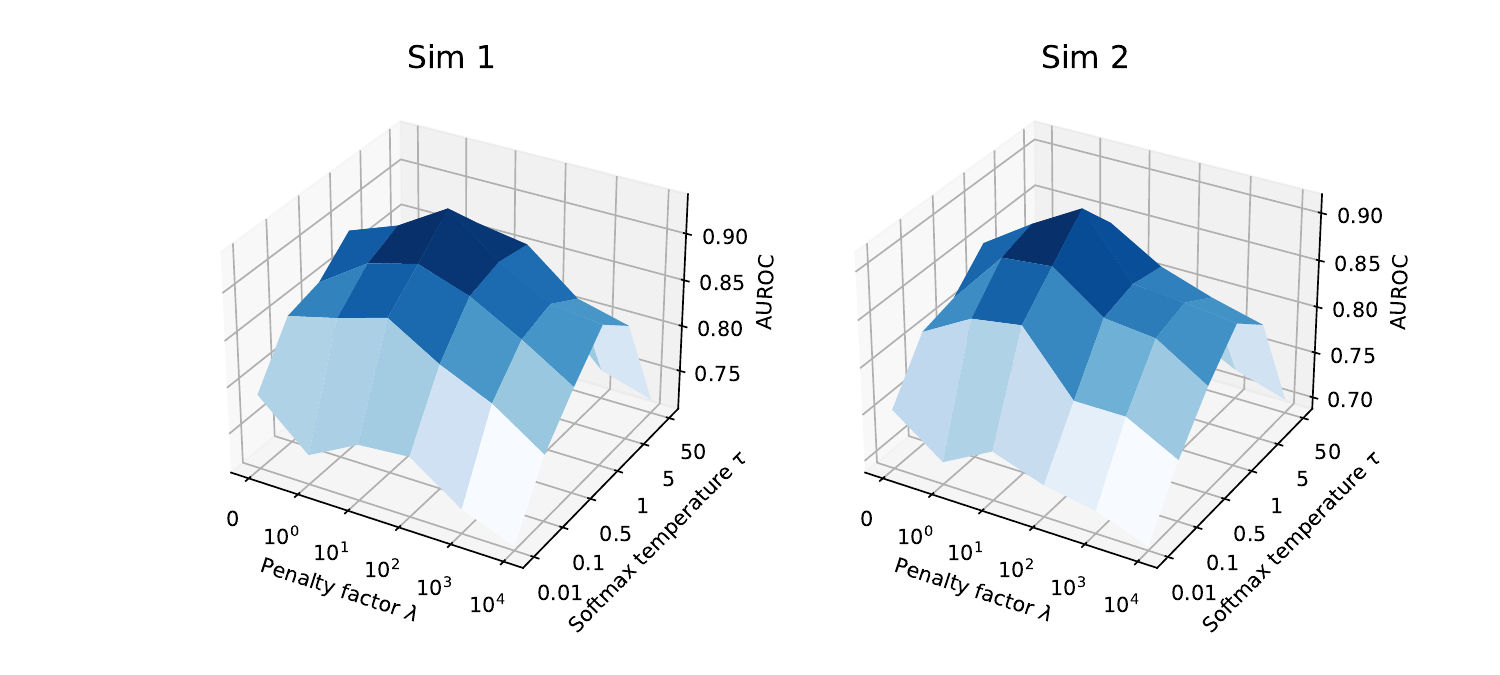} %
	\vspace{-6ex} 
	\caption{Causal structure learning performance \textit{w.r.t.} $\lambda$, $\tau$.} 
	\vspace{-4ex}
	\label{fig:ParaStudy}
\end{figure}

%% file: sections/conclusion.tex
\section{Conclusion}

In this paper, we define the problem of causal MMM, which infers causal structure for each shop and learns the mapping from channel spending to target variable prediction.
To tackle the issues of data heterogeneity and marketing response patterns, we propose \our, which is both provable for learning causal structure from heterogeneous data and capable of modeling patterns in marketing response.
\our employs the causal relational encoder and the marketing response decoder. 
Extensive experiments on the synthetic dataset and real commercial data from an E-commerce platform shows that \our outperforms the baselines and works well in real-world applications.

%% file: sections/acknowledgements.tex
\begin{acks}
    This work has been supported by the National Natural Science Foundation of China under Grant No.: 62002343 and Alibaba Innovative Research (AIR) Program.
    And we would like to express sincere gratitude to the anonymous reviewers for their valuable feedback. 

\end{acks}

%% file: sections/ethical_consideration.tex
\section*{Ethical Considerations}
We conduct causal structure learning from real-world marketing mix data, AirMMM.
Important ethical considerations related to privacy and safety.
Privacy considerations necessitate informed consent, anonymization, and robust data security measures. 
Safety considerations involve responsible dissemination of research findings to avoid unintended negative impacts.
We addressed these considerations and potential misuse concerns with mitigation strategies, including consent, anonymization, transparency, and stakeholder collaboration, to promote responsible research practices.
To be specific, the label of channels are anonymized with only descriptions about media type and means of placement. The shop id and sensitive indicators are processed to avoid information leakage. 
By upholding ethical standards, we can ensure the beneficial and responsible use of online marketing mix data.

%% file: sections/appendix.tex
\appendix

\section{Appendix}
This material serves as the appendix of ``\our: Learning Causal Structures for Marketing Mix Modeling", consisting of theoretical analysis, details of datasets, and further experimental analysis.

\subsection{Theoretical Analysis of \our}

\subsubsection{Granger Causality of \our}\label{app:GC_proof} We prove the following claim to support that we can infer Granger causality in the framework of \our.

\claim{1}{Granger Causality of \our}{For the $n$-th sample in dataset, the variable $\mathbf{x}_{n,i}$ does not Granger cause variable $\mathbf{x}_{n,j}$, if $z_{n,ij}=0$ according to \our.}

To prove the above claim is to show that the j's prediction of $f_{\mathrm{dec}}$ is invariant to $\mathbf{x}_{n,i}^{1:t}$ when \our infers that $\mathbf{x}_{n,i}^{1:t}$ does not cause $\mathbf{x}_{n,i}^{1:t}$.
If the causal relational encoder predicts $z_{n,ij} = 0$ for the nonexistence of edge $i \to j$ on the $n$-th shop, the information of $i$ is blocked in $\mathrm{MSG}_j^t$ according to Equation~\ref{eq_edgeSum}, thus it does not further introduce any new term which depends on $\mathbf{x}_{n,i}$.
Therefore, we can derive that the marketing response decoder's prediction for $j$ is invariant to $\mathbf{x}_{n,i}^{1:t}$ if $z_{n,ij}=0$, and the variable $\mathbf{x}_{n,i}$ does not Granger cause variable $\mathbf{x}_{n,j}$.

\subsubsection{Computational Complexity Analysis}\label{app:complexity_ana}
We discuss the complexity of \our in both training and causal relations inference stages.
For a shop's marketing mix records consisting of $T$ days, we first analyze the time complexity of each module in the training. 
For the causal relational encoder, the time complexity of pairwise embedding and relational interaction is $\mathcal{O}(W)$, where $W$ is the number of weights in neural networks. 
The decoder model marketing response at each time step, so the time complexity is $\mathcal{O}(WT)$.
For inferring causal relations, \our takes marketing mix data as input and calculates structure distributions based on the causal relational encoder. The time complexity is $\mathcal{O}(W)$.
Therefore, the time complexity of the whole model is $\mathcal{O}(WT)$.

\subsection{Details of Datasets}

\subsubsection{Generation of Synthetic Data}\label{app:sim_data}

\begin{figure}[t] 
    \centering
	\includegraphics[width=0.5\textwidth]{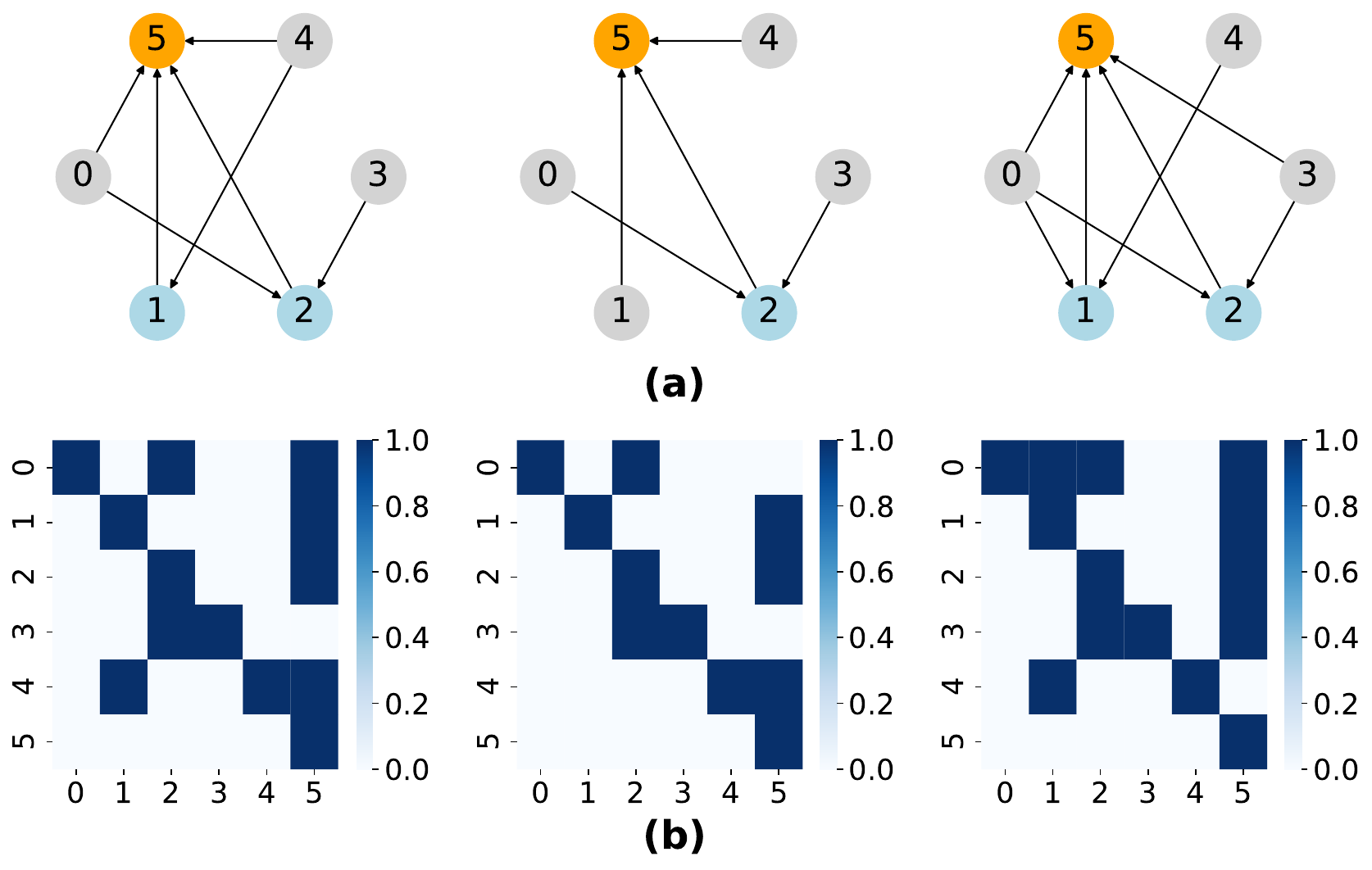} %
	\vspace{-5ex} 
	\caption{(a) Heterogeneous causal structures of synthetic data. (b) The corresponding causal matrix. $A(i,j)=1$ means node $i$ causes node $j$.}
	\vspace{-4ex}
	\label{fig:graph_of_synth}
\end{figure}

We first recap that the number of channel variables is $d$, 
the length of the time series is $T$, and the number of shops is $N$.
We also assume there exist $R$ causal structures in synthetic datasets with $N$ samples.
We aim to obtain temporal data for the marketing mix.
The marketing mix involves two issues~\cite{MMM_interact/vaver2017introduction}: the interactions between different marketing forces and marketing responses.
Therefore, the data generation process is composed of three modules: 
\begin{itemize}
    \item \textbf{Graph sampling.} As illustrated in Figure~\ref{fig:graph_of_synth}, we first generate $R$ directed graphs $\{ \mathcal{G}^1, ..., \mathcal{G}^R\}$ with probability distribution $P_{\mathcal{G}}$, which represents the latent causal structure of marketing variables. 
    For each shop $n$, the latent causal structure $ \mathcal{G}_n$ is sampled from $P_{\mathcal{G}}$. 
    \item \textbf{Channel generation.} Given the latent causal structure $\mathcal{G}_n$ of the $n$-th shop, we can derive the set $\mathcal{S}$ of source nodes, which aren't children of any nodes. We use a similar experimental setup as InGRA~\cite{CD/icdm/InGRA_ChuWMJZY20} to mock temporal data, which follows Non-linear Autoregressive Moving Average (NARMA)~\cite{NARMA/tnn/AtiyaP00} generators:
    \begin{gather*}
        x_{n,i}^t = \alpha_i x_{n,i}^t + \beta_i x_{n,i}^{t-1} \sum_{k=1}^K x_{n,i}^{t-k} + \gamma_i \epsilon_{n}^{t-K}\epsilon_{n}^{t-K} + \epsilon_{n}^{t},  \ ( i \in \mathcal{S})\\
        x_{n, j}^t = \sum_{i=1}^{d} \omega_{n,i} (\eta_{n,j})^{\top} \mathrm{tanh}(\mathbf{x}_{n,i}^{t-K:t-1}) + \epsilon_{n}^{t}, \ ( i \notin \mathcal{S})        \\
    \end{gather*} 
    where $\alpha_i, \beta_i, \gamma_i$ generated from $\mathcal{N}(0,0.1)$ are parameters bounded to marketing variable $m$. 
    $K$ is the order of non-linear interactions, and $\epsilon_{n}^t$ are zero-mean noise terms with $0.01$ variance.
    \item \textbf{Response generation.} We generate the target series via the formula:
    \begin{equation*}  
        u_{n}^t = \sum_{i=1}^{d} \omega_{n,i} (\eta_{n,j})^{\top} \mathrm{tanh}(\mathbf{x}_{n,i}^{t-K:t-1}) + \epsilon_{n}^{t},
    \end{equation*}
The temporal marketing pattern is mocked in the dataset of \textbf{Simulation 1}, where $y_{n}^t = u_{n}^t $.   
Whereas both temporal and saturation marketing patterns are modeled in the dataset of \textbf{Simulation 2}, \ie, $y_{n}^t = f_{curve}(u_{n}^t, c_n)$, and $f_{curve}$ is the S-curve function whose parameters are characterized by $c_n$.
\end{itemize}

\subsubsection{Details of Real-world AirMMM Dataset}\label{app:real_data}

\begin{table*} 
    \centering
    \caption{Descriptions of advertising channels in the real-world AirMMM dataset.}
    \label{tab:airmmm_describe}
     \resizebox{\textwidth}{!}{
        \begin{tabular}{c|c|l}
        \toprule
        Channel Type & Channel Label & Descriptions  \\
        \midrule
        \multirow{6}{*}{brand channel} 
        & x-brand-0 &
        Brand advertising on a video website, including advertising types that appear during video pauses\\
        & x-brand-1 &
        Brand advertising on a video website, including splash screen advertisements\\
        & x-brand-2 &
        Brand advertising on an E-commerce platform\\
        & x-brand-3 &
        Brand advertising which is placed on external websites or platforms\\
        & x-interact &
        Interactive advertising which encourages active engagement and participation from the users\\
        & x-feed &
        Feed advertising which appears within social media feeds or content feeds on target platforms\\
        \midrule
        \multirow{5}{*}{effect channel} 
        & x-live &
        Advertising takes place during live events on an E-commerce platform\\
        & x-search-0 &
        Search advertising placed on external websites or platforms\\
        & x-search-1 &
        Search advertising on an E-commerce platform\\
        & x-search-2 &
        Search advertising on an E-commerce platform\\
        & x-effect &
        Effect advertising\\
        \midrule
        \multirow{2}{*}{others} 
        & PV &
        Page views \\
        & GMV &
        Gross merchandise value\\
        \bottomrule
        \end{tabular}}
        \vspace{-4ex}

\end{table*}

Collected from a representative E-commerce platform, the AirMMM dataset contains 50 shops' marketing mix data with a period of $22$ months from Jan 30th 2021 to Dec 6th 2022.
It consists of costs on $11$ advertising channels, PV (page views, which can also be viewed as channels in the causal graph), and GMV, as described in Table~\ref{tab:airmmm_describe}.

These channels can be categorized into two groups, \ie, brand channels and effect channels. 
Brand channels are designed to capture users' interest and promote awareness of the target brand, including x-brand-0/1/2/3, x-interact, and x-feed. 
In contrast, X-live, x-search-0/1/2, and x-effect are effect channels where users are more likely to take action and convert. These channels in various forms (interaction, content feed, live, and \etc) are placed on different platforms such as video websites, E-commerce platforms, and news browsers.

\subsection{Further Experimental Analysis}\label{app:experimental}

\subsubsection{Training Curve}

\begin{figure}[h] 
    \centering
	\includegraphics[width=0.47\textwidth]{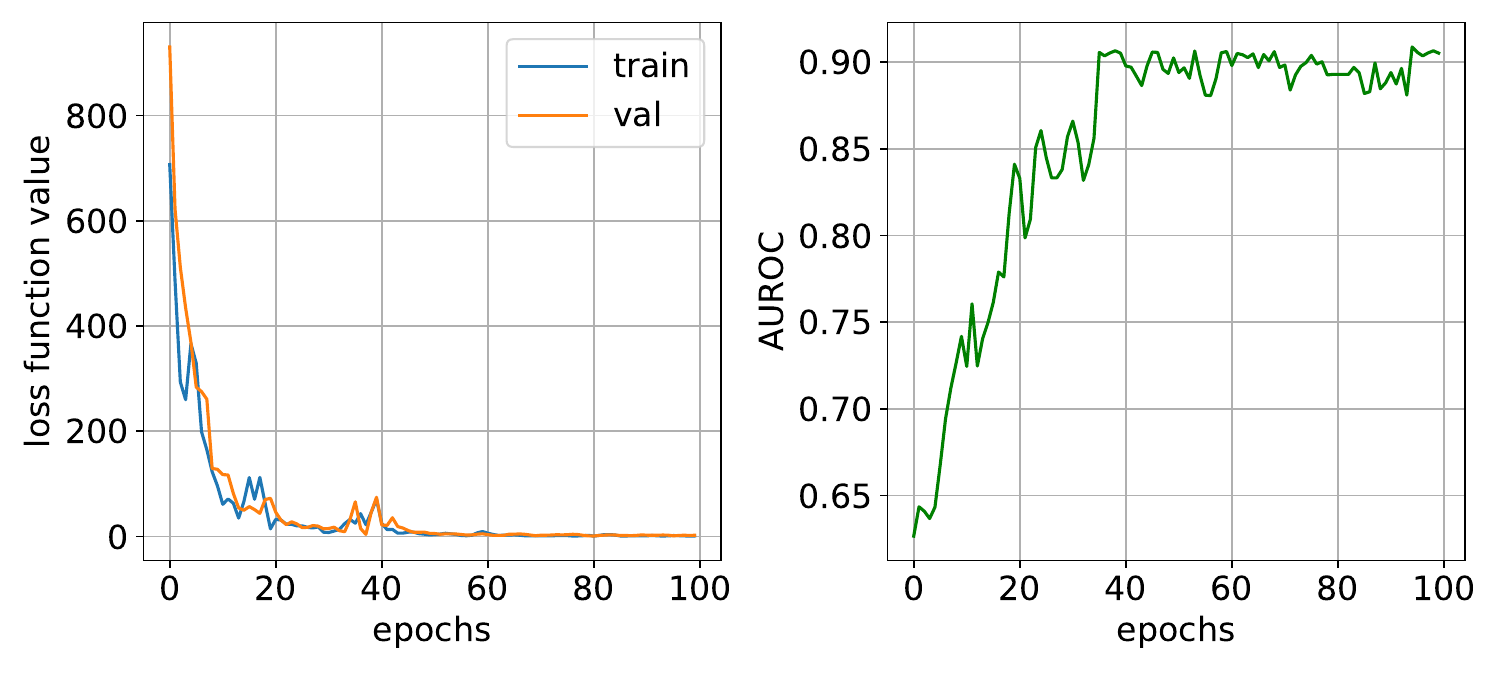} %
	\vspace{-3ex}
	\caption{Training, validation loss, and AUROC across epochs.} %
	\vspace{-2ex}
	\label{fig:LOSS_AUROC_EPOCH_PLOT}
\end{figure}

To show the convergence of our \our method, we plot the training loss, validation loss, and AUROC metric across training epochs in Figure~\ref{fig:LOSS_AUROC_EPOCH_PLOT}.
Firstly, it's evident that the training and validation losses converge well, demonstrating that \our is reasonably stable.
Secondly, it shows that AUROC rises simultaneously when loss converges, which illustrates that the quality of the causal structure learning improves as the error of marketing response modeling converges.

\subsubsection{Comparison of Training Scalability}\label{app:running_time}
\begin{figure}[h] 
    \centering
	\includegraphics[width=0.4\textwidth]{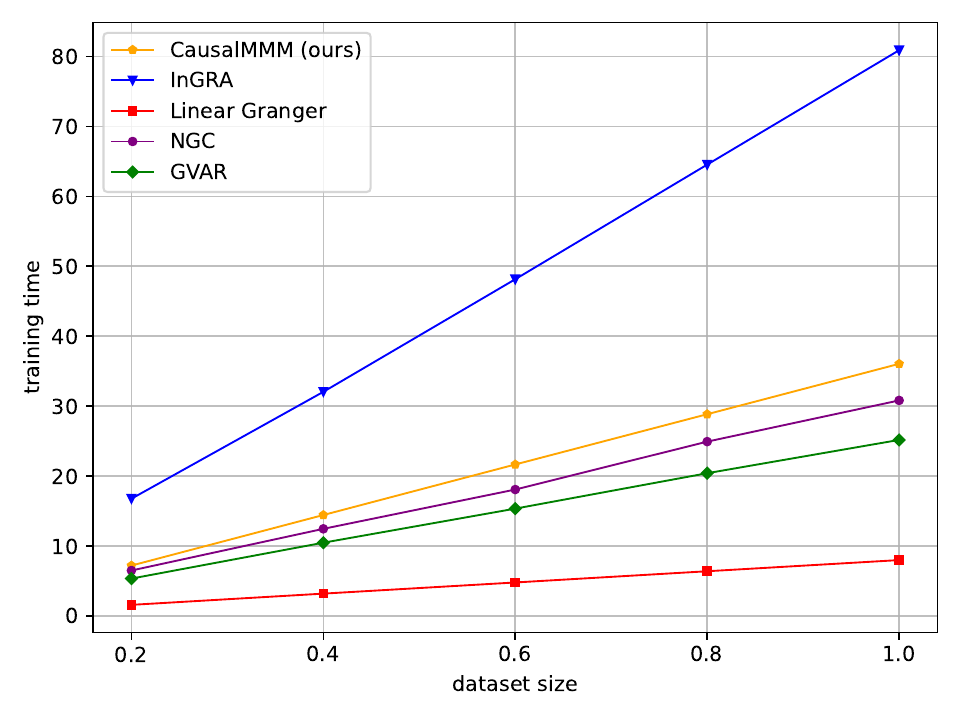} 
	\vspace{-4ex} 
	\caption{Comparison of training scalability.} 
	\label{fig:training_scalability}
\end{figure}

We investigate the training scalability of \our, InGRA, Linear Granger, NGC, and GVAR.
We vary the size of the training dataset from $20\%$ to $100\%$ to simulate the increase of online shops to be served.
The results of Linear Granger, NGC, and GVAR are evaluated by summing training time on each shop.
We train all models on a high-end server with $2\times$ NVIDIA GTX 3090 GPUs. 
The results are shown in Figure~\ref{fig:training_scalability}.

From the figure, we can observe that all methods scale linearly with the increase of the training data size. 
Therefore, they're promising to handle massive marketing data from a large number of online shops.
Analogous to \our, InGRA can utilize information across different shops.  
However, its time consumption is much higher than the proposed \our because of the computational complexity of InGRA's attention mechanism.

%% file: main.bbl

\begin{thebibliography}{45}


\ifx \showCODEN    \undefined \def \showCODEN     #1{\unskip}     \fi
\ifx \showDOI      \undefined \def \showDOI       #1{#1}\fi
\ifx \showISBNx    \undefined \def \showISBNx     #1{\unskip}     \fi
\ifx \showISBNxiii \undefined \def \showISBNxiii  #1{\unskip}     \fi
\ifx \showISSN     \undefined \def \showISSN      #1{\unskip}     \fi
\ifx \showLCCN     \undefined \def \showLCCN      #1{\unskip}     \fi
\ifx \shownote     \undefined \def \shownote      #1{#1}          \fi
\ifx \showarticletitle \undefined \def \showarticletitle #1{#1}   \fi
\ifx \showURL      \undefined \def \showURL       {\relax}        \fi
\providecommand\bibfield[2]{#2}
\providecommand\bibinfo[2]{#2}
\providecommand\natexlab[1]{#1}
\providecommand\showeprint[2][]{arXiv:#2}

\bibitem[Arabzadeh et~al\mbox{.}(2018)]%
        {DBLP:conf/cikm/ArabzadehFZNB18}
\bibfield{author}{\bibinfo{person}{Negar Arabzadeh}, \bibinfo{person}{Hossein Fani}, \bibinfo{person}{Fattane Zarrinkalam}, \bibinfo{person}{Ahmed Navivala}, {and} \bibinfo{person}{Ebrahim Bagheri}.} \bibinfo{year}{2018}\natexlab{}.
\newblock \showarticletitle{Causal Dependencies for Future Interest Prediction on Twitter}. In \bibinfo{booktitle}{\emph{Proceedings of the 27th {ACM} International Conference on Information and Knowledge Management, {CIKM} 2018, Torino, Italy, October 22-26, 2018}}. \bibinfo{publisher}{{ACM}}, \bibinfo{pages}{1511--1514}.
\newblock


\bibitem[Arnold et~al\mbox{.}(2007)]%
        {kdd/ArnoldLA07}
\bibfield{author}{\bibinfo{person}{Andrew Arnold}, \bibinfo{person}{Yan Liu}, {and} \bibinfo{person}{Naoki Abe}.} \bibinfo{year}{2007}\natexlab{}.
\newblock \showarticletitle{Temporal causal modeling with graphical granger methods}. In \bibinfo{booktitle}{\emph{Proceedings of the 13th {ACM} {SIGKDD} International Conference on Knowledge Discovery and Data Mining, San Jose, California, USA, August 12-15, 2007}}. \bibinfo{publisher}{{ACM}}, \bibinfo{pages}{66--75}.
\newblock


\bibitem[Assaad et~al\mbox{.}(2022)]%
        {intro/ts_surveys/AssaadDG22}
\bibfield{author}{\bibinfo{person}{Charles~K. Assaad}, \bibinfo{person}{Emilie Devijver}, {and} \bibinfo{person}{{\'{E}}ric Gaussier}.} \bibinfo{year}{2022}\natexlab{}.
\newblock \showarticletitle{Survey and Evaluation of Causal Discovery Methods for Time Series}.
\newblock \bibinfo{journal}{\emph{J. Artif. Intell. Res.}}  \bibinfo{volume}{73} (\bibinfo{year}{2022}), \bibinfo{pages}{767--819}.
\newblock
\urldef\tempurl%
\url{https://doi.org/10.1613/jair.1.13428}
\showDOI{\tempurl}


\bibitem[Atiya and Parlos(2000)]%
        {NARMA/tnn/AtiyaP00}
\bibfield{author}{\bibinfo{person}{Amir~F. Atiya} {and} \bibinfo{person}{Alexander~G. Parlos}.} \bibinfo{year}{2000}\natexlab{}.
\newblock \showarticletitle{New results on recurrent network training: unifying the algorithms and accelerating convergence}.
\newblock \bibinfo{journal}{\emph{{IEEE} Trans. Neural Networks Learn. Syst.}} \bibinfo{volume}{11}, \bibinfo{number}{3} (\bibinfo{year}{2000}), \bibinfo{pages}{697--709}.
\newblock


\bibitem[Banijamali(2022)]%
        {NRI/iclr/Banijamali22}
\bibfield{author}{\bibinfo{person}{Ershad Banijamali}.} \bibinfo{year}{2022}\natexlab{}.
\newblock \showarticletitle{Neural Relational Inference with Node-Specific Information}. In \bibinfo{booktitle}{\emph{The Tenth International Conference on Learning Representations, {ICLR} 2022, Virtual Event, April 25-29, 2022}}.
\newblock


\bibitem[Cai et~al\mbox{.}(2023)]%
        {DBLP:conf/wsdm/CaiFWYLX23}
\bibfield{author}{\bibinfo{person}{Wei Cai}, \bibinfo{person}{Fuli Feng}, \bibinfo{person}{Qifan Wang}, \bibinfo{person}{Tian Yang}, \bibinfo{person}{Zhenguang Liu}, {and} \bibinfo{person}{Congfu Xu}.} \bibinfo{year}{2023}\natexlab{}.
\newblock \showarticletitle{A Causal View for Item-level Effect of Recommendation on User Preference}. In \bibinfo{booktitle}{\emph{Proceedings of the Sixteenth {ACM} International Conference on Web Search and Data Mining, {WSDM} 2023, Singapore, 27 February 2023 - 3 March 2023}}. \bibinfo{publisher}{{ACM}}, \bibinfo{pages}{240--248}.
\newblock


\bibitem[Chan and Perry(2017)]%
        {MMM_google/chan2017challenges}
\bibfield{author}{\bibinfo{person}{David Chan} {and} \bibinfo{person}{Mike Perry}.} \bibinfo{year}{2017}\natexlab{}.
\newblock \showarticletitle{Challenges and opportunities in media mix modeling}.
\newblock  (\bibinfo{year}{2017}).
\newblock


\bibitem[Chen et~al\mbox{.}(2018)]%
        {MMM_google/chen2018bias}
\bibfield{author}{\bibinfo{person}{Aiyou Chen}, \bibinfo{person}{David Chan}, \bibinfo{person}{Mike Perry}, \bibinfo{person}{Yuxue Jin}, \bibinfo{person}{Yunting Sun}, \bibinfo{person}{Yueqing Wang}, {and} \bibinfo{person}{Jim Koehler}.} \bibinfo{year}{2018}\natexlab{}.
\newblock \showarticletitle{Bias correction for paid search in media mix modeling}.
\newblock \bibinfo{journal}{\emph{arXiv preprint arXiv:1807.03292}} (\bibinfo{year}{2018}).
\newblock


\bibitem[Chen et~al\mbox{.}(2021)]%
        {JAS/chen2021hierarchical}
\bibfield{author}{\bibinfo{person}{Hao Chen}, \bibinfo{person}{Minguang Zhang}, \bibinfo{person}{Lanshan Han}, {and} \bibinfo{person}{Alvin Lim}.} \bibinfo{year}{2021}\natexlab{}.
\newblock \showarticletitle{Hierarchical marketing mix models with sign constraints}.
\newblock \bibinfo{journal}{\emph{Journal of Applied Statistics}} \bibinfo{volume}{48}, \bibinfo{number}{13-15} (\bibinfo{year}{2021}), \bibinfo{pages}{2944--2960}.
\newblock


\bibitem[Cheng et~al\mbox{.}(2016)]%
        {forWideDeep/recsys/Cheng0HSCAACCIA16}
\bibfield{author}{\bibinfo{person}{Heng{-}Tze Cheng}, \bibinfo{person}{Levent Koc}, \bibinfo{person}{Jeremiah Harmsen}, \bibinfo{person}{Tal Shaked}, \bibinfo{person}{Tushar Chandra}, \bibinfo{person}{Hrishi Aradhye}, \bibinfo{person}{Glen Anderson}, \bibinfo{person}{Greg Corrado}, \bibinfo{person}{Wei Chai}, \bibinfo{person}{Mustafa Ispir}, \bibinfo{person}{Rohan Anil}, \bibinfo{person}{Zakaria Haque}, \bibinfo{person}{Lichan Hong}, \bibinfo{person}{Vihan Jain}, \bibinfo{person}{Xiaobing Liu}, {and} \bibinfo{person}{Hemal Shah}.} \bibinfo{year}{2016}\natexlab{}.
\newblock \showarticletitle{Wide {\&} Deep Learning for Recommender Systems}. In \bibinfo{booktitle}{\emph{Proceedings of the 1st Workshop on Deep Learning for Recommender Systems, DLRS@RecSys 2016, Boston, MA, USA, September 15, 2016}}. \bibinfo{publisher}{{ACM}}, \bibinfo{pages}{7--10}.
\newblock


\bibitem[Cheng et~al\mbox{.}(2022)]%
        {DBLP:conf/wsdm/ChengG022}
\bibfield{author}{\bibinfo{person}{Lu Cheng}, \bibinfo{person}{Ruocheng Guo}, {and} \bibinfo{person}{Huan Liu}.} \bibinfo{year}{2022}\natexlab{}.
\newblock \showarticletitle{Estimating Causal Effects of Multi-Aspect Online Reviews with Multi-Modal Proxies}. In \bibinfo{booktitle}{\emph{{WSDM} '22: The Fifteenth {ACM} International Conference on Web Search and Data Mining, Virtual Event / Tempe, AZ, USA, February 21 - 25, 2022}}. \bibinfo{publisher}{{ACM}}, \bibinfo{pages}{103--112}.
\newblock


\bibitem[Chu et~al\mbox{.}(2020)]%
        {CD/icdm/InGRA_ChuWMJZY20}
\bibfield{author}{\bibinfo{person}{Yunfei Chu}, \bibinfo{person}{Xiaowei Wang}, \bibinfo{person}{Jianxin Ma}, \bibinfo{person}{Kunyang Jia}, \bibinfo{person}{Jingren Zhou}, {and} \bibinfo{person}{Hongxia Yang}.} \bibinfo{year}{2020}\natexlab{}.
\newblock \showarticletitle{Inductive Granger Causal Modeling for Multivariate Time Series}. In \bibinfo{booktitle}{\emph{20th {IEEE} International Conference on Data Mining, {ICDM} 2020, Sorrento, Italy, November 17-20, 2020}}. \bibinfo{pages}{972--977}.
\newblock


\bibitem[Gerhardus and Runge(2020)]%
        {MTS/CB/LPCMCI_GerhardusR20}
\bibfield{author}{\bibinfo{person}{Andreas Gerhardus} {and} \bibinfo{person}{Jakob Runge}.} \bibinfo{year}{2020}\natexlab{}.
\newblock \showarticletitle{High-recall causal discovery for autocorrelated time series with latent confounders}. In \bibinfo{booktitle}{\emph{Advances in Neural Information Processing Systems 33: Annual Conference on Neural Information Processing Systems 2020, NeurIPS 2020, December 6-12, 2020, virtual}}.
\newblock


\bibitem[Gong et~al\mbox{.}(2023)]%
        {KDD_tutorial_ours}
\bibfield{author}{\bibinfo{person}{Chang Gong}, \bibinfo{person}{Di Yao}, \bibinfo{person}{Chuzhe Zhang}, \bibinfo{person}{Wenbin Li}, \bibinfo{person}{Jingping Bi}, \bibinfo{person}{Lun Du}, {and} \bibinfo{person}{Jin Wang}.} \bibinfo{year}{2023}\natexlab{}.
\newblock \showarticletitle{Causal Discovery from Temporal Data}. In \bibinfo{booktitle}{\emph{Proceedings of the 29th ACM SIGKDD Conference on Knowledge Discovery and Data Mining}} \emph{(\bibinfo{series}{KDD '23})}. \bibinfo{pages}{5803–5804}.
\newblock


\bibitem[Granger(1969)]%
        {granger1969investigating}
\bibfield{author}{\bibinfo{person}{Clive~WJ Granger}.} \bibinfo{year}{1969}\natexlab{}.
\newblock \showarticletitle{Investigating causal relations by econometric models and cross-spectral methods}.
\newblock \bibinfo{journal}{\emph{Econometrica: journal of the Econometric Society}} (\bibinfo{year}{1969}), \bibinfo{pages}{424--438}.
\newblock


\bibitem[Hanssens et~al\mbox{.}(2003)]%
        {bg_adeffect/hanssens2003market}
\bibfield{author}{\bibinfo{person}{Dominique~M Hanssens}, \bibinfo{person}{Leonard~J Parsons}, {and} \bibinfo{person}{Randall~L Schultz}.} \bibinfo{year}{2003}\natexlab{}.
\newblock \bibinfo{booktitle}{\emph{Market response models: Econometric and time series analysis}}. Vol.~\bibinfo{volume}{2}.
\newblock \bibinfo{publisher}{Springer Science \& Business Media}.
\newblock


\bibitem[Hochreiter and Schmidhuber(1997)]%
        {forLSTM/neco/HochreiterS97}
\bibfield{author}{\bibinfo{person}{Sepp Hochreiter} {and} \bibinfo{person}{J{\"{u}}rgen Schmidhuber}.} \bibinfo{year}{1997}\natexlab{}.
\newblock \showarticletitle{Long Short-Term Memory}.
\newblock \bibinfo{journal}{\emph{Neural Comput.}} \bibinfo{volume}{9}, \bibinfo{number}{8} (\bibinfo{year}{1997}), \bibinfo{pages}{1735--1780}.
\newblock


\bibitem[Jang et~al\mbox{.}(2017)]%
        {Gumbel/iclr/JangGP17}
\bibfield{author}{\bibinfo{person}{Eric Jang}, \bibinfo{person}{Shixiang Gu}, {and} \bibinfo{person}{Ben Poole}.} \bibinfo{year}{2017}\natexlab{}.
\newblock \showarticletitle{Categorical Reparameterization with Gumbel-Softmax}. In \bibinfo{booktitle}{\emph{5th International Conference on Learning Representations, {ICLR} 2017, Toulon, France, April 24-26, 2017, Conference Track Proceedings}}.
\newblock


\bibitem[Jin et~al\mbox{.}(2017)]%
        {MMM_46001/jin2017bayesian}
\bibfield{author}{\bibinfo{person}{Yuxue Jin}, \bibinfo{person}{Yueqing Wang}, \bibinfo{person}{Yunting Sun}, \bibinfo{person}{David Chan}, {and} \bibinfo{person}{Jim Koehler}.} \bibinfo{year}{2017}\natexlab{}.
\newblock \showarticletitle{Bayesian methods for media mix modeling with carryover and shape effects}.
\newblock  (\bibinfo{year}{2017}).
\newblock


\bibitem[Khanna and Tan(2020)]%
        {MTS/Granger/iclr20_esru}
\bibfield{author}{\bibinfo{person}{Saurabh Khanna} {and} \bibinfo{person}{Vincent Y.~F. Tan}.} \bibinfo{year}{2020}\natexlab{}.
\newblock \showarticletitle{Economy Statistical Recurrent Units For Inferring Nonlinear Granger Causality}. In \bibinfo{booktitle}{\emph{8th International Conference on Learning Representations, {ICLR} 2020, Addis Ababa, Ethiopia, April 26-30, 2020}}.
\newblock


\bibitem[Kipf et~al\mbox{.}(2018)]%
        {NRI/icml/KipfFWWZ18}
\bibfield{author}{\bibinfo{person}{Thomas~N. Kipf}, \bibinfo{person}{Ethan Fetaya}, \bibinfo{person}{Kuan{-}Chieh Wang}, \bibinfo{person}{Max Welling}, {and} \bibinfo{person}{Richard~S. Zemel}.} \bibinfo{year}{2018}\natexlab{}.
\newblock \showarticletitle{Neural Relational Inference for Interacting Systems}. In \bibinfo{booktitle}{\emph{Proceedings of the 35th International Conference on Machine Learning, {ICML} 2018, Stockholmsm{\"{a}}ssan, Stockholm, Sweden, July 10-15, 2018}} \emph{(\bibinfo{series}{Proceedings of Machine Learning Research}, Vol.~\bibinfo{volume}{80})}. \bibinfo{publisher}{{PMLR}}, \bibinfo{pages}{2693--2702}.
\newblock


\bibitem[L{\"{o}}we et~al\mbox{.}(2022)]%
        {Discussion/NewForm/ACD_LoweMSW22}
\bibfield{author}{\bibinfo{person}{Sindy L{\"{o}}we}, \bibinfo{person}{David Madras}, \bibinfo{person}{Richard~Z. Shilling}, {and} \bibinfo{person}{Max Welling}.} \bibinfo{year}{2022}\natexlab{}.
\newblock \showarticletitle{Amortized Causal Discovery: Learning to Infer Causal Graphs from Time-Series Data}. In \bibinfo{booktitle}{\emph{1st Conference on Causal Learning and Reasoning, CLeaR 2022, Sequoia Conference Center, Eureka, CA, USA, 11-13 April, 2022}} \emph{(\bibinfo{series}{Proceedings of Machine Learning Research}, Vol.~\bibinfo{volume}{177})}. \bibinfo{publisher}{{PMLR}}, \bibinfo{pages}{509--525}.
\newblock


\bibitem[Malinsky and Spirtes(2018)]%
        {MTS/CB_FCI_SVAR_FCI_MalinskyS18}
\bibfield{author}{\bibinfo{person}{Daniel Malinsky} {and} \bibinfo{person}{Peter Spirtes}.} \bibinfo{year}{2018}\natexlab{}.
\newblock \showarticletitle{Causal Structure Learning from Multivariate Time Series in Settings with Unmeasured Confounding}. In \bibinfo{booktitle}{\emph{Proceedings of 2018 {ACM} {SIGKDD} Workshop on Causal Discovery, CD@KDD 2018, London, UK, 20 August 2018}} \emph{(\bibinfo{series}{Proceedings of Machine Learning Research}, Vol.~\bibinfo{volume}{92})}. \bibinfo{publisher}{{PMLR}}, \bibinfo{pages}{23--47}.
\newblock


\bibitem[Marcinkevics and Vogt(2021)]%
        {MTS/Granger/iclr21_GVAR_MarcinkevicsV}
\bibfield{author}{\bibinfo{person}{Ricards Marcinkevics} {and} \bibinfo{person}{Julia~E. Vogt}.} \bibinfo{year}{2021}\natexlab{}.
\newblock \showarticletitle{Interpretable Models for Granger Causality Using Self-explaining Neural Networks}. In \bibinfo{booktitle}{\emph{9th International Conference on Learning Representations, {ICLR} 2021, Virtual Event, Austria, May 3-7, 2021}}.
\newblock


\bibitem[Mart{\'\i}nez et~al\mbox{.}(2022)]%
        {bg_adeffect/martinez2022distributed}
\bibfield{author}{\bibinfo{person}{Andr{\'e}s Mart{\'\i}nez}, \bibinfo{person}{Alfonso Salafranca}, \bibinfo{person}{Ana~E Sipols}, \bibinfo{person}{Clara~Simon de Blas}, {and} \bibinfo{person}{Daniel van Hengel}.} \bibinfo{year}{2022}\natexlab{}.
\newblock \showarticletitle{Distributed lags using elastic-net regularization for market response models: focus on predictive and explanatory capacity}.
\newblock \bibinfo{journal}{\emph{Journal of Marketing Analytics}} (\bibinfo{year}{2022}), \bibinfo{pages}{1--19}.
\newblock


\bibitem[Moraffah et~al\mbox{.}(2021)]%
        {intro/ts_surveys/Moraffah21}
\bibfield{author}{\bibinfo{person}{Raha Moraffah}, \bibinfo{person}{Paras Sheth}, \bibinfo{person}{Mansooreh Karami}, \bibinfo{person}{Anchit Bhattacharya}, \bibinfo{person}{Qianru Wang}, \bibinfo{person}{Anique Tahir}, \bibinfo{person}{Adrienne Raglin}, {and} \bibinfo{person}{Huan Liu}.} \bibinfo{year}{2021}\natexlab{}.
\newblock \showarticletitle{Causal inference for time series analysis: problems, methods and evaluation}.
\newblock \bibinfo{journal}{\emph{Knowl. Inf. Syst.}} \bibinfo{volume}{63}, \bibinfo{number}{12} (\bibinfo{year}{2021}), \bibinfo{pages}{3041--3085}.
\newblock


\bibitem[Ng et~al\mbox{.}(2021)]%
        {adkdd/ng2021bayesian}
\bibfield{author}{\bibinfo{person}{Edwin Ng}, \bibinfo{person}{Zhishi Wang}, {and} \bibinfo{person}{Athena Dai}.} \bibinfo{year}{2021}\natexlab{}.
\newblock \showarticletitle{Bayesian Time Varying Coefficient Model with Applications to Marketing Mix Modeling}. In \bibinfo{booktitle}{\emph{Proceedings of AdKDD 2021 Workshop}}.
\newblock


\bibitem[Nuara et~al\mbox{.}(2019)]%
        {NN_bg_causal/www/NuaraSTZ0R19}
\bibfield{author}{\bibinfo{person}{Alessandro Nuara}, \bibinfo{person}{Nicola Sosio}, \bibinfo{person}{Francesco Trov{\`{o}}}, \bibinfo{person}{Maria~Chiara Zaccardi}, \bibinfo{person}{Nicola Gatti}, {and} \bibinfo{person}{Marcello Restelli}.} \bibinfo{year}{2019}\natexlab{}.
\newblock \showarticletitle{Dealing with Interdependencies and Uncertainty in Multi-Channel Advertising Campaigns Optimization}. In \bibinfo{booktitle}{\emph{The World Wide Web Conference, {WWW} 2019, San Francisco, CA, USA, May 13-17, 2019}}. \bibinfo{pages}{1376--1386}.
\newblock


\bibitem[Pamfil et~al\mbox{.}(2020)]%
        {MTS/SB/Dynotears_aistats_PamfilSDPGBA20}
\bibfield{author}{\bibinfo{person}{Roxana Pamfil}, \bibinfo{person}{Nisara Sriwattanaworachai}, \bibinfo{person}{Shaan Desai}, \bibinfo{person}{Philip Pilgerstorfer}, \bibinfo{person}{Konstantinos Georgatzis}, \bibinfo{person}{Paul Beaumont}, {and} \bibinfo{person}{Bryon Aragam}.} \bibinfo{year}{2020}\natexlab{}.
\newblock \showarticletitle{{DYNOTEARS:} Structure Learning from Time-Series Data}. In \bibinfo{booktitle}{\emph{The 23rd International Conference on Artificial Intelligence and Statistics, {AISTATS} 2020, 26-28 August 2020, Online [Palermo, Sicily, Italy]}} \emph{(\bibinfo{series}{Proceedings of Machine Learning Research}, Vol.~\bibinfo{volume}{108})}. \bibinfo{publisher}{{PMLR}}, \bibinfo{pages}{1595--1605}.
\newblock


\bibitem[Pearl(2009)]%
        {pearl2009causality}
\bibfield{author}{\bibinfo{person}{Judea Pearl}.} \bibinfo{year}{2009}\natexlab{}.
\newblock \bibinfo{booktitle}{\emph{Causality}}.
\newblock \bibinfo{publisher}{Cambridge university press}.
\newblock


\bibitem[Peters et~al\mbox{.}(2013)]%
        {MTS/FCM/nips_PetersJS13}
\bibfield{author}{\bibinfo{person}{Jonas Peters}, \bibinfo{person}{Dominik Janzing}, {and} \bibinfo{person}{Bernhard Sch{\"{o}}lkopf}.} \bibinfo{year}{2013}\natexlab{}.
\newblock \showarticletitle{Causal Inference on Time Series using Restricted Structural Equation Models}. In \bibinfo{booktitle}{\emph{Advances in Neural Information Processing Systems 26: 27th Annual Conference on Neural Information Processing Systems 2013. Proceedings of a meeting held December 5-8, 2013, Lake Tahoe, Nevada, United States}}. \bibinfo{pages}{154--162}.
\newblock


\bibitem[Shojaie and Fox(2021)]%
        {Granger_review}
\bibfield{author}{\bibinfo{person}{Ali Shojaie} {and} \bibinfo{person}{Emily~B. Fox}.} \bibinfo{year}{2021}\natexlab{}.
\newblock \showarticletitle{Granger Causality: {A} Review and Recent Advances}.
\newblock \bibinfo{journal}{\emph{CoRR}}  \bibinfo{volume}{abs/2105.02675} (\bibinfo{year}{2021}).
\newblock
\showeprint[arXiv]{2105.02675}
\urldef\tempurl%
\url{https://arxiv.org/abs/2105.02675}
\showURL{%
\tempurl}


\bibitem[Sun et~al\mbox{.}(2021)]%
        {MTS/SB/NTS_NOTEARS}
\bibfield{author}{\bibinfo{person}{Xiangyu Sun}, \bibinfo{person}{Guiliang Liu}, \bibinfo{person}{Pascal Poupart}, {and} \bibinfo{person}{Oliver Schulte}.} \bibinfo{year}{2021}\natexlab{}.
\newblock \showarticletitle{{NTS-NOTEARS:} Learning Nonparametric Temporal DAGs With Time-Series Data and Prior Knowledge}.
\newblock \bibinfo{journal}{\emph{CoRR}}  \bibinfo{volume}{abs/2109.04286} (\bibinfo{year}{2021}).
\newblock


\bibitem[Sun et~al\mbox{.}(2017)]%
        {MMM_google/sun2017geo}
\bibfield{author}{\bibinfo{person}{Yunting Sun}, \bibinfo{person}{Yueqing Wang}, \bibinfo{person}{Yuxue Jin}, \bibinfo{person}{David Chan}, {and} \bibinfo{person}{Jim Koehler}.} \bibinfo{year}{2017}\natexlab{}.
\newblock \showarticletitle{Geo-level bayesian hierarchical media mix modeling}.
\newblock  (\bibinfo{year}{2017}).
\newblock


\bibitem[Tan et~al\mbox{.}(2022)]%
        {DBLP:conf/wsdm/TanZHKYYZYPZ022}
\bibfield{author}{\bibinfo{person}{Ziqi Tan}, \bibinfo{person}{Shengyu Zhang}, \bibinfo{person}{Nuanxin Hong}, \bibinfo{person}{Kun Kuang}, \bibinfo{person}{Yifan Yu}, \bibinfo{person}{Jin Yu}, \bibinfo{person}{Zhou Zhao}, \bibinfo{person}{Hongxia Yang}, \bibinfo{person}{Shiyuan Pan}, \bibinfo{person}{Jingren Zhou}, {and} \bibinfo{person}{Fei Wu}.} \bibinfo{year}{2022}\natexlab{}.
\newblock \showarticletitle{Uncovering Causal Effects of Online Short Videos on Consumer Behaviors}. In \bibinfo{booktitle}{\emph{{WSDM} '22: The Fifteenth {ACM} International Conference on Web Search and Data Mining, Virtual Event / Tempe, AZ, USA, February 21 - 25, 2022}}. \bibinfo{publisher}{{ACM}}, \bibinfo{pages}{997--1006}.
\newblock


\bibitem[Tank et~al\mbox{.}(2022)]%
        {MTS/Granger/pamiNGC22}
\bibfield{author}{\bibinfo{person}{Alex Tank}, \bibinfo{person}{Ian Covert}, \bibinfo{person}{Nicholas~J. Foti}, \bibinfo{person}{Ali Shojaie}, {and} \bibinfo{person}{Emily~B. Fox}.} \bibinfo{year}{2022}\natexlab{}.
\newblock \showarticletitle{Neural Granger Causality}.
\newblock \bibinfo{journal}{\emph{{IEEE} Trans. Pattern Anal. Mach. Intell.}} \bibinfo{volume}{44}, \bibinfo{number}{8} (\bibinfo{year}{2022}), \bibinfo{pages}{4267--4279}.
\newblock


\bibitem[Vaver and Zhang(2017)]%
        {MMM_interact/vaver2017introduction}
\bibfield{author}{\bibinfo{person}{Jon Vaver} {and} \bibinfo{person}{Stephanie Shin-Hui Zhang}.} \bibinfo{year}{2017}\natexlab{}.
\newblock \showarticletitle{Introduction to the Aggregate Marketing System Simulator}.
\newblock  (\bibinfo{year}{2017}).
\newblock


\bibitem[Vowels et~al\mbox{.}(2023)]%
        {csur/VowelsCB23}
\bibfield{author}{\bibinfo{person}{Matthew~J. Vowels}, \bibinfo{person}{Necati~Cihan Camg{\"{o}}z}, {and} \bibinfo{person}{Richard Bowden}.} \bibinfo{year}{2023}\natexlab{}.
\newblock \showarticletitle{D'ya Like DAGs? {A} Survey on Structure Learning and Causal Discovery}.
\newblock \bibinfo{journal}{\emph{{ACM} Comput. Surv.}} \bibinfo{volume}{55}, \bibinfo{number}{4} (\bibinfo{year}{2023}), \bibinfo{pages}{82:1--82:36}.
\newblock


\bibitem[Wang et~al\mbox{.}(2017)]%
        {MMM_google/wang2017hierarchical}
\bibfield{author}{\bibinfo{person}{Yueqing Wang}, \bibinfo{person}{Yuxue Jin}, \bibinfo{person}{Yunting Sun}, \bibinfo{person}{David Chan}, {and} \bibinfo{person}{Jim Koehler}.} \bibinfo{year}{2017}\natexlab{}.
\newblock \showarticletitle{A hierarchical Bayesian approach to improve media mix models using category data}.
\newblock  (\bibinfo{year}{2017}).
\newblock


\bibitem[Weichwald et~al\mbox{.}(2019)]%
        {nips/WeichwaldJMPTV19}
\bibfield{author}{\bibinfo{person}{Sebastian Weichwald}, \bibinfo{person}{Martin~Emil Jakobsen}, \bibinfo{person}{Phillip~B. Mogensen}, \bibinfo{person}{Lasse Petersen}, \bibinfo{person}{Nikolaj Thams}, {and} \bibinfo{person}{Gherardo Varando}.} \bibinfo{year}{2019}\natexlab{}.
\newblock \showarticletitle{Causal structure learning from time series: Large regression coefficients may predict causal links better in practice than small p-values}. In \bibinfo{booktitle}{\emph{NeurIPS 2019 Competition and Demonstration Track, 8-14 December 2019, Vancouver, Canada. Revised selected papers}} \emph{(\bibinfo{series}{Proceedings of Machine Learning Research}, Vol.~\bibinfo{volume}{123})}. \bibinfo{publisher}{{PMLR}}, \bibinfo{pages}{27--36}.
\newblock


\bibitem[Wigren and Cornell(2019)]%
        {comp_survey/wigren2019marketing}
\bibfield{author}{\bibinfo{person}{Richard Wigren} {and} \bibinfo{person}{Filip Cornell}.} \bibinfo{year}{2019}\natexlab{}.
\newblock \bibinfo{title}{Marketing Mix Modelling: A comparative study of statistical models}.
\newblock
\newblock


\bibitem[Wu et~al\mbox{.}(2020)]%
        {MPIR/tailin}
\bibfield{author}{\bibinfo{person}{Tailin Wu}, \bibinfo{person}{Thomas~M. Breuel}, \bibinfo{person}{Michael Skuhersky}, {and} \bibinfo{person}{Jan Kautz}.} \bibinfo{year}{2020}\natexlab{}.
\newblock \showarticletitle{Discovering Nonlinear Relations with Minimum Predictive Information Regularization}.
\newblock \bibinfo{journal}{\emph{CoRR}}  \bibinfo{volume}{abs/2001.01885} (\bibinfo{year}{2020}).
\newblock
\urldef\tempurl%
\url{http://arxiv.org/abs/2001.01885}
\showURL{%
\tempurl}


\bibitem[Wu et~al\mbox{.}(2022)]%
        {MTS/FCM/cikm_WuWWLC22}
\bibfield{author}{\bibinfo{person}{Tianhao Wu}, \bibinfo{person}{Xingyu Wu}, \bibinfo{person}{Xin Wang}, \bibinfo{person}{Shikang Liu}, {and} \bibinfo{person}{Huanhuan Chen}.} \bibinfo{year}{2022}\natexlab{}.
\newblock \showarticletitle{Nonlinear Causal Discovery in Time Series}. In \bibinfo{booktitle}{\emph{Proceedings of the 31st {ACM} International Conference on Information {\&} Knowledge Management, Atlanta, GA, USA, October 17-21, 2022}}. \bibinfo{publisher}{{ACM}}, \bibinfo{pages}{4575--4579}.
\newblock


\bibitem[Xu et~al\mbox{.}(2019)]%
        {MTS/Granger/SCGL_CIKM_XuHY19}
\bibfield{author}{\bibinfo{person}{Chenxiao Xu}, \bibinfo{person}{Hao Huang}, {and} \bibinfo{person}{Shinjae Yoo}.} \bibinfo{year}{2019}\natexlab{}.
\newblock \showarticletitle{Scalable Causal Graph Learning through a Deep Neural Network}. In \bibinfo{booktitle}{\emph{Proceedings of the 28th {ACM} International Conference on Information and Knowledge Management, {CIKM} 2019, Beijing, China, November 3-7, 2019}}. \bibinfo{publisher}{{ACM}}, \bibinfo{pages}{1853--1862}.
\newblock


\bibitem[Yao et~al\mbox{.}(2022)]%
        {NN_bg/kdd/YaoGZCB22}
\bibfield{author}{\bibinfo{person}{Di Yao}, \bibinfo{person}{Chang Gong}, \bibinfo{person}{Lei Zhang}, \bibinfo{person}{Sheng Chen}, {and} \bibinfo{person}{Jingping Bi}.} \bibinfo{year}{2022}\natexlab{}.
\newblock \showarticletitle{CausalMTA: Eliminating the User Confounding Bias for Causal Multi-touch Attribution}. In \bibinfo{booktitle}{\emph{{KDD} '22: The 28th {ACM} {SIGKDD} Conference on Knowledge Discovery and Data Mining, Washington, DC, USA, August 14 - 18, 2022}}. \bibinfo{pages}{4342--4352}.
\newblock


\end{thebibliography}
